%% file: main.tex
\documentclass{interact} 

\usepackage{graphicx}
\usepackage{url}
\usepackage{multirow}
\usepackage{epstopdf}
\usepackage[caption=false]{subfig}
\usepackage{wrapfig}
\usepackage{caption}
\usepackage{svg}
\usepackage{lmodern}
\usepackage[numbers,sort&compress]{natbib}
\bibpunct[, ]{[}{]}{,}{n}{,}{,}
\renewcommand\bibfont{\fontsize{10}{12}\selectfont}
\makeatletter
\def\NAT@def@citea{\def\@citea{\NAT@separator}}
\makeatother

\theoremstyle{plain}

\theoremstyle{definition}

\theoremstyle{remark}

\DeclareMathOperator*{\argmax}{arg\,max}

\begin{document}

\title{Social Comparison without Explicit Inference of Others' Reward Values: A Constructive Approach Using a Probabilistic Generative Model}

\author{
    \name{
        Yosuke Taniuchi\textsuperscript{a}\thanks{CONTACT Yosuke Taniuchi. Email: taniuchi.yosuke.ub9@is.naist.jp},
        Chie Hieida\textsuperscript{a}, Atsushi Noritake\textsuperscript{b, c, d}, 
        Kazushi Ikeda\textsuperscript{a} and Masaki Isoda\textsuperscript{c,d}
    }
    \affil{
        \textsuperscript{a} Division of Information Science, Graduate  School  of  Science  and  Technology, Nara Institute of Science and Technology, Ikoma, Japan
        \textsuperscript{b} Department of Integrative Neuroscience, Graduate School of Medicine and Pharmaceutical Sciences, University of Toyama, Toyama, Japan
        \textsuperscript{c} Division of Behavioral Development, Department of System Neuroscience, National Institute for Physiological Sciences, National Institutes of Natural Sciences, Okazaki, Japan
        \textsuperscript{d} Department of Physiological Sciences, School of Life Science, The Graduate University for Advanced Studies (SOKENDAI), Hayama, Japan
    }
}

\thanks{This work was supported by JSPS KAKENHI, 22H05082, 24K00515, 22H05081, 22H04931, 24K17239, 25H00579, Japan.}
\thanks{This work has been submitted to the IEEE for possible publication. Copyright may be transferred without notice, after which this version may no longer be accessible.}

\maketitle
\input{sections/abstract}


\input{sections/introduction}
\input{sections/materials_and_methods}
\input{sections/results}

\input{sections/discussion}
\input{sections/conclusion}

\section*{Acknowledgments}
We would like to express our gratitude to Professor Takayuki Nagai and Professor Kazuhiro Tanaka for their valuable discussions. 

\section*{Disclosure statement}
No potential conflict of interest was reported by the author(s).

\section*{Additional information}
\subsection*{Funding}
This work was supported by JSPS KAKENHI, 22H05082, 24K00515, 22H05081, 22H04931, 24K17239, 25H00579, Japan.

\bibliographystyle{tfnlm}
\bibliography{references} 

\newpage
\include{sections/appendix}
\end{document}

%% file: sections/abstract.tex
\begin{abstract}
Social comparison---the process of evaluating one's rewards relative to others---is an essential feature of social emotions such as envy and plays a fundamental role in primate social cognition. However, it remains unknown how information about others' rewards affects one's own reward valuation. This study examines whether monkeys merely recognize objective differences in reward or instead infer others' subjective reward valuations. To address this issue, a constructive approach---one that replicates target emotions in artificial systems and extracts knowledge from them---was employed, owing to its potential to simulate how the monkey interacts with social contexts, specifically social comparison. We developed three computational models with varying degrees of social information processing: an Internal Prediction Model (IPM), which infers the partner's subjective values; a No Comparison Model (NCM), which disregards partner information; and an External Comparison Model (ECM), which directly incorporates the partner's objective rewards. We trained the models on a dataset containing the behavior of a pair of monkeys, their rewards, and the conditioned stimuli, and then evaluated the models' ability to classify subjective values across pre-defined experimental conditions. The ECM achieved the best classification result (0.88 for the ECM vs. 0.85 for the IPM on the Rand index), suggesting that, in our modeling framework, social comparison relies on objective differences in reward rather than on inferences about subjective reward values.
\end{abstract}

%% file: sections/introduction.tex
\section{Introduction}
Envy is an emotion typically defined as the `pain at another's good fortune~\cite{Tai2012-pk}. It is fundamentally linked to social comparison---the comparison of one's rewards with those of others~\cite{festinger1954theory}. It can lead to hostility~\cite{Tai2012-pk} and poorer mental health~\cite{Castillo-Gualda2026-ex}, but it may also promote productivity~\cite{Tai2012-pk}. Thus, the effects of envy vary across situations. However, the mechanisms underlying envy remain poorly understood.  
Understanding these mechanisms is important for predicting and controlling the effects of envy in daily life.

A constructive approach offers a promising framework for investigating the mechanisms of envy, particularly in affective robotics and affective computing. It can replicate cognitive and emotional processes, including envy, by creating complete artificial systems, such as robots or computational models. Their designs and parameters within this framework may provide insight into the mechanisms of envy, as the replication process is expected to enable the system to encode information about envy~\cite{Scheier2021-nx}. Indeed, prior constructive research has contributed substantially to our understanding of the mechanisms underlying basic emotions~\cite{Hieida2022-ns, Asada2009-wj}, including how bodily perception, linguistic information, and visual stimuli are integrated into emotional categories~\cite {Tsurumaki2025-ai}. However, constructive studies of emotion in social contexts, particularly social comparison, remain largely unexplored regarding envy. A major obstacle is the lack of datasets that capture social comparison processes. 

An interesting biological study on social comparison by Noritake et al.~\cite{Noritake2018-hm}), from which this study gained its dataset, revealed that social comparison occurs not only in humans but also in non-human primates: macaque monkeys exhibit behavioral markers of social comparison, such as changes in anticipatory licking, and neural correlates in midbrain dopaminergic neurons~\cite{Noritake2018-hm}. A key finding of social comparison in monkeys is that while objective rewards remain constant, the subjective value that monkeys assign can vary depending on the context~\cite{Noritake2018-hm}. In their experiments, monkeys repeatedly received rewards in the presence of a partner. The subjective value of rewards declined as the partner's reward probability increased, although the objective number of outcomes remained constant for the self-monkey. This shift was confirmed neurally and behaviorally through altered activation of midbrain dopaminergic neurons and corresponding changes in anticipatory behavior~\cite{Noritake2018-hm}.

The experiment by Noritake et al.~\cite{Noritake2018-hm} demonstrated subjective valuation in monkeys, and the result makes us curious not only about the subjective value for oneself but also their inference of \emph{the partner's subjective value}. Given recent evidence that monkeys exhibit theory of mind abilities---attributing beliefs distinct from their own and predicting others' actions~\cite{Hayashi2020-qk, Drayton2016-yv}, we are prompted to investigate whether monkeys can infer their partner's reward valuations.
If monkeys can represent others' mental states, they may infer the value a partner assigns to a reward and incorporate this inference into their own valuation during social comparison. Another study of theory of mind in non-human primates, including macaques~\cite{devaine2017reading} showed that non-human primates engaged in theory of mind inference during gameplay related to partner valuation, but not during social comparison. Thus, the study motivates questioning whether the subjective value inference about the partner applies to social comparison settings.

Despite extensive research on social comparison in primates, few studies have directly examined whether humans or non-human primates understand others' subjective reward valuations. Significant progress---e.g., by Noritake et al.~\cite{Noritake2018-hm, Noritake2020-os}---has clarified how primates monitor and evaluate their own rewards relative to others, yet evidence for representations of others' subjective valuations remains scarce. 
Despite the lack of research, two prior computational studies have explored related questions about partner valuation problems, though they address different topics than social comparison~\cite{Zaki2016-vr, devaine2017reading}.
Zaki et al.~\cite{Zaki2016-vr} combined fMRI with reinforcement learning models to demonstrate that humans can estimate others' emotional states, which may involve the other's subjective valuation. Their paradigm, however, centered on empathic emotion inference rather than on changes in valuation during social comparison. 
Devaine et al.~\cite{devaine2017reading} investigated non-human primates' (including macaques) theory of mind ability in terms of how they infer their partner's actions in competitive games, which are also different from social comparison.
Therefore, neither study directly examined how an agent's inference about a partner's subjective value influences its own valuation during social comparison, which is the focus of this study.
Consequently, the computational principles linking a represented other-value to one's own \emph{subjective value} remain largely unexplored.

Given the limited research on inferring others' subjective value in social comparison settings, the present study asks: when a monkey compares its reward with a partner's, is its subjective value influenced by an inference about the partner's subjective valuation, or is it determined solely by the objective reward difference? 

The limited investigation of this question reflects the inherently analytic focus of most neuroscientific approaches, which isolate specific neural or behavioral activities from complex cognitive systems. As shown by Noritake et al.~\cite{Noritake2018-hm, Noritake2020-os}, these studies examined causal interactions among brain regions---the medial prefrontal cortex (MPFC), lateral hypothalamus, and dopaminergic midbrain nuclei.
Although such analytic approaches reveal necessary neural conditions for social comparison, these components alone constitute a sufficient cognitive mechanism that is deployable in robotic systems.

This study adopts a constructive approach that complements analytical methods by building and analyzing artificial systems to clarify a comprehensive understanding of the target mechanisms~\cite{Asada2001-jq, Taniguchi2024-wg}. Using computational model selection with behavioral and perceptual data from primates, we identify frameworks that best explain observed patterns. This approach is particularly effective for examining internal representations---such as how inferences about others' valuations shape one's subjective value---when direct neural recordings are unavailable. 

To compare social comparison models with and without causal links from inferred partner's subjective values to one's own, we employed the multi-layered, multimodal Latent Dirichlet Allocation (mMLDA) ~\cite{Nakamura2009-dv}, which is an extension of the classical LDA~\cite{Blei2003-wg}.
mMLDA was applied to extract emotion categories from multimodal data while retaining an interpretable latent structure~\cite{Tsurumaki2025-ai}, because the problem is essentially similar to mMLDA's typical problems of categorizing multimodal inputs without predefined labels~\cite{Taniguchi2016-dx}.
Due to its interpretability and multimodality, mMLDA is well-suited for modeling how monkeys may internally represent subjective reward values in social contexts involving multimodal data about the self and the partner. Reinforcement learning is not a good candidate because the dataset lacks feedback loops and does not require action-policy optimization.
We experimentally evaluated the fit of each model across six conditions, using the dataset collected by Noritake et al.~\cite{Noritake2018-hm}. We leveraged mMLDA's categorization capacity, with a latent node to estimate subjective reward valuations.

%% file: sections/materials_and_methods.tex
\section{Materials and Methods}
We aimed to determine whether monkeys infer their partner's subjective reward values or respond solely to objective rewards.
To investigate this, we developed and compared three probabilistic generative models with varying capacities for processing social information:
I) Internal Prediction Model (IPM), which attempts to infer the partner's subjective valuations;
II) No Comparison Model (NCM), which processes only self-information; and 
III) External Comparison Model (ECM), which directly incorporates observed rewards received by the partner.
We evaluated the extent to which each model captured subjective-value representations across the six experimental conditions.

To implement these models, we used mMLDA~\cite{Attamimi2014-yw}, which enabled modeling of subjective reward values as latent variables while simultaneously processing multimodal inputs, including behavioral and reward data from both the self and partner monkeys, as well as stimulus images. The mMLDA is a hierarchical composition of MLDA~\cite{Nakamura2009-dv}, with its technical details provided in Appendix~\ref{app:mlda}. All models were implemented using the SERKET framework~\cite{Nakamura2018-rg}. The SERKET framework connects programmatically independent modules to construct cognitive models. By employing a message passing approach, SERKET integrates multiple LDA and MLDA inferences to achieve mMLDA-based inference. 

\subsection{Behavioral Tasks and Dataset}
\label{dataset}
\begin{figure*}[!t]
    \centering
    \includegraphics[width=0.6\textwidth]{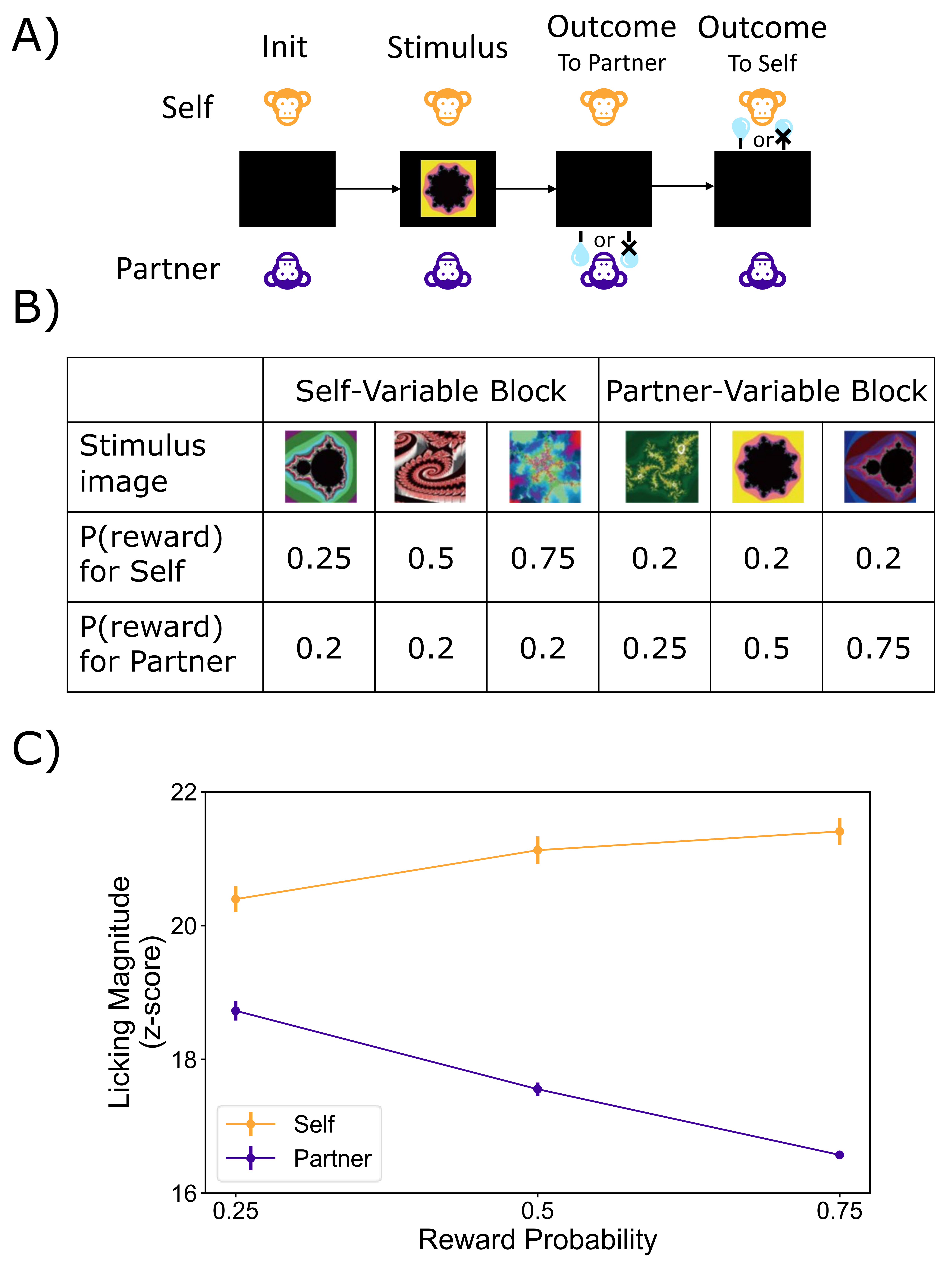}
    \caption{
        \textbf{A)} Experimental setup: two face-to-face monkeys separated by a horizontally placed LCD monitor, which showed one of the six stimulus images. 
        The water nozzles, equipped with sensors to measure anticipatory licking, were placed beside the monkeys.
        \textbf{B)} Experimental conditions: The self-variable blocks (left) with variable reward probabilities (25\%, 50\%, or 75\%) for the self monkey and fixed 20\% for the partner; the partner-variable blocks (right) with variable probabilities for the partner and fixed 20\% for the self monkey.
        \textbf{C)} Results showing decreased licking when the partner received higher rewards (purple), despite the self monkey's fixed 20\% reward probability, and increased licking as the self monkey's own reward probability increased (yellow). Center and error bars indicate mean $\pm$ s.e.m.
    }
    \label{fig:experimental_settings}
\end{figure*}
We used the dataset collected in the experiment by Noritake et al.~\cite{Noritake2018-hm}, which involved two monkeys in a social Pavlovian conditioning setup over 292 days, during which partner-licking data and self-licking data were recorded.
One monkey was designated the `self' and the other the `partner'.
During each trial, the monkeys faced each other across a horizontally placed LCD monitor between them, which displayed a stimulus image, and each monkey's anticipatory licking behavior was recorded as a measure of reward expectation (Figure~\ref{fig:experimental_settings}A).
The experiment featured two types of blocks, each with three levels of reward probability, yielding six distinct experimental conditions.
In each condition, a unique visual stimulus was displayed on the monitor. In the self-variable blocks, the self monkey received rewards with variable probabilities (25\%, 50\%, or 75\%), while the partner received rewards with a fixed probability of 20\%. In the partner-variable blocks, the partner received rewards with variable probabilities (25\%, 50\%, or 75\%), while the self monkey received rewards with a fixed 20\% probability (Figure~\ref{fig:experimental_settings}B).
In the experiments, monkeys exhibited anticipatory licking responses that predicted upcoming rewards, and the frequency of these responses correlated with dopaminergic neuronal activity. Therefore, we can reasonably interpret different licking frequency patterns as indicating various levels of subjective reward value~\cite{Noritake2018-hm}.
Anticipatory licking behavior was analyzed using the same time windows as found in the study by Noritake et al.~\cite{Noritake2018-hm} (see Appendix~\ref{app:licking}).
For training our computational models, a random 80--20 split of the 292-day dataset was used, with 233 days for training and 59 days for testing, ensuring generalization and validation on data not included in the training phase. Each day's data was divided into six experimental blocks, each treated as a single `document' in the terminology of the LDA. Details of feature vector extraction from these multimodal data and the composition of the documents are provided in Appendix~\ref{app:feature}.

\subsection{Model Architectures}

\subsubsection{Basic Structures}
\begin{table}
    \caption{Notations for five modal observations and latent variables}
    \centering
    \begin{tabular}{|l|l|}
        \hline
        \multicolumn{2}{|l|}{Observation} \\
        \hline
        Notation & Meaning\\
        \hline
        $w^{As}/w^{Ap}$ & Licking frequency (Self/ Partner)\\
        $w^{Rs}/w^{Rp}$ & Reward frequency (Self/ Partner)\\
        $w^{S}$  & Visual stimuli\\
        \hline
        \hline
        \multicolumn{2}{|l|}{Latent Variable} \\
        \hline
        Notation & Meaning\\
        \hline
        $z^{As}/z^{Ap}$ & Filtered licking frequency (Self/ Partner)\\
        $z^{Rs}/z^{Rp}$ & Self subjective valuation (Self/ Partner)\\
        $z^S$    & Situation awareness\\
        \hline
    \end{tabular}
    \label{tab:model_notation}
\end{table}
To introduce our model structures, we initially extract the essential observation of social comparison from the first-person perspective of the self-monkey in the experiments by Noritake et al.~\cite{Noritake2018-hm}. The self-monkey first recognized the visual stimuli ($w^S$) and then perceived whether the partner received water as a reward ($w^{Rp}$). Next, the self-monkey received or did not receive a reward ($w^{Rs}$). Simultaneously, the self-monkey could see how frequently the partner monkey licks the nozzle in anticipation of rewards, which was recorded. The self-monkey's anticipatory licking behavior was recorded in parallel. According to Noritake et al.~\cite{Noritake2018-hm}, within the time window between the presentation of visual stimuli and the presentation of partner reward, the self-monkey's licking frequency and the activation of dopaminergic neurons of the midbrain significantly correlated, signifying that the self licking frequency at this time period ($w^{As}$) reflected the subjective valuation of the self monkey ($z^{Rs}$). The same time window was applied to extract the partner licking frequency ($w^{Ap}$), which should reflect the partner's subjective valuation ($z^{Rp}$). 
The notations for the observations and latent variables are summarized in Table~\ref{tab:model_notation}, though some of the variables ($z^S, z^{A*}$) will be introduced below. 

\begin{figure*}
    \centering
    \includegraphics[width=0.6\linewidth]{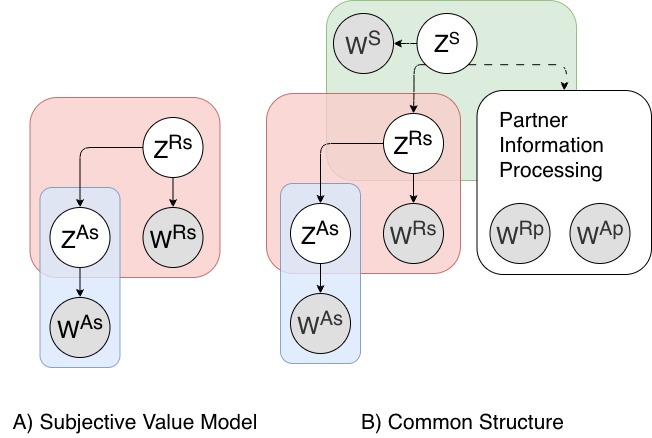}
    \caption{A) Subjective Value Model: a building block for our more complex architectures in terms of self-subjective valuation. B) Common Structure of our models that added the visual stimuli and situation awareness variable. For simplicity, this figure shows only the observation and latent variables.}
    \label{fig:subjective_value_model}
\end{figure*}

Before designing the full generative process of social comparison with the complex multimodal data described above, we focus here on the generative process of self-information (licking $w^{As}$ and reward $w^{Rs}$). As latent variables, an action intention ($z^{As}$) and subjective valuation ($z^{Rs}$) were added. We call this theoretical building block the Subjective Value Model.
First, the connection between self-licking frequency and subjective valuation should be determined by the finding by Noritake et al.~\cite{Noritake2018-hm}: the self-licking frequency in the time window could be interpreted to be correlated to the self-subjective valuation based on Noritake et al.~\cite{Noritake2018-hm}. Therefore, self-licking frequency was hypothesized to be caused by the latent variable of self-subjective valuation ($z^{Rs}\rightarrow w^{As}$, we use `$\rightarrow$' for generative relationships). Second, a filtered licking ($z^{As}$) was added in the middle to reduce the inherent noise of the biological data ($z^{Rs}\rightarrow z^{As} \rightarrow w^{As}$), which represented the self-monkey's action intention. Lastly, the self-reward frequency was designed to be predicted from the self-subjective valuation ($z^{Rs} \rightarrow w^{Rs}$), based on the temporal order ($z^{Rs}$ before $w^{As}$). Thus, the self-subjective valuation learns the co-occurrence between self-licking and reward. The resulting generative process of the Subjective Value Model is summarized in the Figure~\ref{fig:subjective_value_model} A.

To consider how the visual stimuli ($w^{S}$) and the partner information ($w^{*p}$) have a causal influence on the Subjective Value Model, a situation awareness variable ($z^S$) was introduced to the Subjective Value Model. After repetitive training of the experimental situations by Noritake et al.~\cite{Noritake2018-hm}, the self-monkey should learn to categorize the co-occurrence of the visual stimuli ($w^{S}$) and the subsequent observations of each monkey ($w^{*s}, w^{*p}$). This recognition in the self-monkey's mind is labeled as situation awareness ($z^S$). By separating the partner's generative process from the Subjective Value Model and retaining its specific structure, which we describe as the three variants in the next subsection, we obtain the Common Structure shown in Figure~\ref{fig:subjective_value_model}B. The Common Structure posits that situation awareness generates self and partner information separately, with the former generated by the Subjective Value Model. 

\subsubsection{The Individual Model Architecture}
Based on the Subjective Value Model and Common Structure, three variants of mMLDA models were constructed depending on the varieties of partner information processing:

\begin{enumerate}
    \item[I)] \textbf{Internal Prediction Model (IPM)}: assumes that monkeys infer their partner's subjective reward valuations.
    \item[II)] \textbf{No Comparison Model (NCM)}: assumes monkeys process only their own rewards without incorporating social information.
    \item[III)] \textbf{External Comparison Model (ECM)}: assumes monkeys recognize their partner's rewards but without inferring subjective valuations.
\end{enumerate}

\begin{figure*}[!t]
    \centering
    \includegraphics[width=\textwidth]{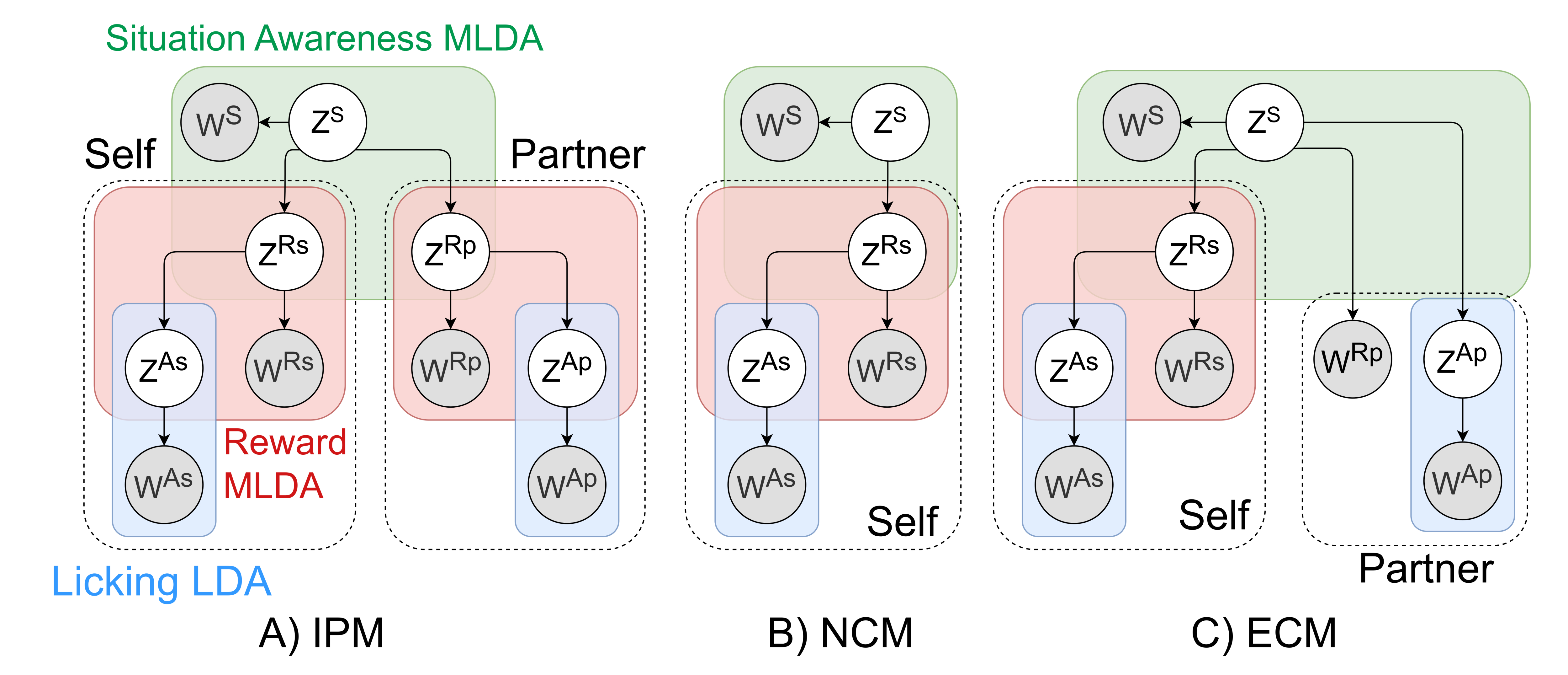}
    \caption{
        Comparison of model architectures:
        A) IPM: incorporating inference of the partner's subjective values. 
        B) NCM: processing only self-related information.
        C) ECM: including the partner's observed rewards but not their subjective valuation.
        For simplicity, the generative parameters $\theta^*$ and $\phi^*$ and hyperparameters $\alpha^*$ and $\beta^*$ are omitted, showing only the relationships between observed variables $w^*$ and latent variables $z^*$.
        }
    \label{fig:model_architectures}
\end{figure*}

The IPM assumes that the self-monkey processes partner-related information the same way it processes self information. In other words, not only the self's subjective valuation but also the partner's subjective valuation is generated or inferred from the self-monkey's perspective. Therefore, the Subjective Value Model of the self is replicated as the partner processing model, and the IPM is distinguished from the other two by a latent variable for the partner's subjective valuation ($z^{Rp}$) within the partner's Subjective Value Model (Figure~\ref{fig:model_architectures}A). 
Because of the symmetry, the model can also be interpreted as a model of experimental situations from the third-person perspective. But the entire model is intended to describe the essential cognitive process inside the monkey's brain during social comparison. Note that, in reality, interoception, or observations accessible only to oneself, such as heartbeat, should differentiate self- and partner-processing, but because we focus on behavior and objective rewards, nothing distinguishes the self-processing model from the partner-processing model.

The NCM, as a control model, omits all partner-related information ($w^{Ap}, w^{Rp}, z^{Ap},z^{Rp}$). And the situation awareness latent variable ($z^S$) only generates the image observation ($w^s$) and the subjective value of self-reward ($z^{Rs}$), as shown in Figure~\ref{fig:model_architectures}B.

The ECM posited that the subjective value of the partner reward is not inferred with the latent variable representing it ($z^{Rp}$) omitted. Instead, the model assumed that the situation awareness variable ($z^S$) directly predicts the partner reward observation and the filtered licking frequency ($z^{Ap}$). By contrasting the IPM and ECM, we sought to examine whether the self-monkey predicted the partner's subjective valuations during social comparison. The ECM's structure is described in Figure~\ref{fig:model_architectures}C.

To compare these three model structures, the IPM has the largest parameter count due to the partner valuation inference variable, suggesting potentially better performance on the metrics introduced in the next subsection. On the other hand, the ECM lacked that parameter while maintaining its partner information-processing abilities. If the ECM performs better on the metrics, the variable used for partner valuation inference can reasonably be considered redundant.
Because it cannot process partner information, the NCM is not supposed to perform well and should serve only as a control model. 

The figures for the three models are simplified for readability, and the exact Bayesian formulations of the models and mathematical formulations of the model training and prediction process are described in Appendix~\ref{app:math_bayesian}. The determination of the models' hyperparameters is provided in Appendix~\ref{app:hyper}.

\subsection{Model Validation}
Model performance was evaluated primarily by the classification accuracy for the self-monkey \emph{subjective-value} variable, allowing us to test whether the IPM discriminated subjective values more precisely than simpler baselines.

Three complementary analyses were performed on the held-out test data. 
\begin{enumerate}
    \item[I)] We assessed how accurately each model categorized the self-monkey's subjective value variable. The six experimental conditions were treated as six levels of subjective value---an assumption supported by condition-specific correlations between dopaminergic activity and anticipatory licking~\cite{Noritake2018-hm}. A model that assigns trials to the correct condition with higher accuracy was therefore considered a better description of the monkeys' valuation process.
    \item[II)] We tested whether each model reproduced the self monkey's anticipatory licking behavior. We evaluated \emph{interpolation} and \emph{extrapolation} in this regard. For interpolation, the models were given only the visual cues and predicted licking. Each model was expected to generate the same trends as found in Figure~\ref{fig:experimental_settings}C. For extrapolation, reward distributions missing in the training set---such as a \(65\,\%\) self-reward versus \(20\,\%\) partner-reward configuration (see Figure~\ref{fig:experimental_settings}B)---were supplied, and the predicted licking was evaluated against empirical data.
    \item[III)] We quantified the influence of every variable on the self-subjective-value variable by computing the \emph{normalized mutual information} between them. This analysis revealed how the flow of information into the self-valuation variable varied across model architectures. This index is widely used, for example, in community detection~\cite{Girvan2002-dp}, feature selection~\cite{Estevez2009-lr}, and as a regularizer in variational auto-encoders~\cite{Zhao2017-hr}. Here, it provides a principled, scale-free metric for highlighting which pathways in the multimodal generative models carried the greatest share of task-relevant information. The details of the computation of normalized mutual information are provided in Appendix~\ref{app:nmi}.
\end{enumerate}

%% file: sections/results.tex
\section{Results}
\subsection{Classification Performance of Self-Valuation Node}
\subsubsection{Evaluation with Rand Index}
\begin{table}[t] 
    \centering
    \caption{Classification Accuracy of Subjective-Value Nodes}
    \begin{tabular}{lcccc}
    \hline
    \textbf{Metric} & \textbf{IPM} & \textbf{NCM} & \textbf{ECM} & \textbf{Chance Level} \\
    \hline
    Rand Index & 0.85 & 0.83 & 0.88 & 0.72 \\
    Adjusted Rand Index & 0.56 & 0.57 & 0.60 & - \\
    \hline
    \end{tabular}
    \label{table:classification_accuracy}
\end{table}
The ECM demonstrated the highest classification accuracy for subjective-value categorization. We used the Rand index~\cite{Hubert1985-wh} and adjusted Rand index~\cite{Hubert1985-wh} as the performance metrics, both of which measure the similarity between two clusterings by calculating the proportion of trial pairs that were either grouped or separated in both the model's clustering and the actual experimental conditions. The difference between the Rand index and the adjusted Rand index is that the latter is corrected for chance. The two metrics were adopted as complementary: the Rand index was difficult to interpret due to its high chance level in our setting, whereas the adjusted Rand index was unintuitive because it does not directly show the correct grouping ratios. As of the Rand index, against the chance level of 0.72 (for how to gain the chance level, see Appendi~\ref{app:chance_level}), the ECM achieved an accuracy of 0.88, outperforming both the NCM (0.83) and IPM (0.85) (see Table~\ref{table:classification_accuracy}). Regarding the adjusted Rand index, the ECM also ranked first with an accuracy of 0.60, and, in this metric, the NCM (0.57) was better than the IPM (0.56). For the reason for the different rankings between the Rand index and the adjusted Rand index, see Appendix~\ref{app:adjusted_rand_index}.

The superior performance of the ECM suggests that direct observation of objective reward differences is more suitable for categorizing subjective values than inferring the partner's subjective experience.

\subsubsection{Clustering Analysis of Subjective Value Representation}
To better understand the effectiveness of each model's architecture for representing subjective value, we conducted a detailed analysis of clustering patterns in the latent space of the self-subjective-value topic vectors.

\begin{figure*}[t!]
\centering
\includegraphics[width=0.7\linewidth]{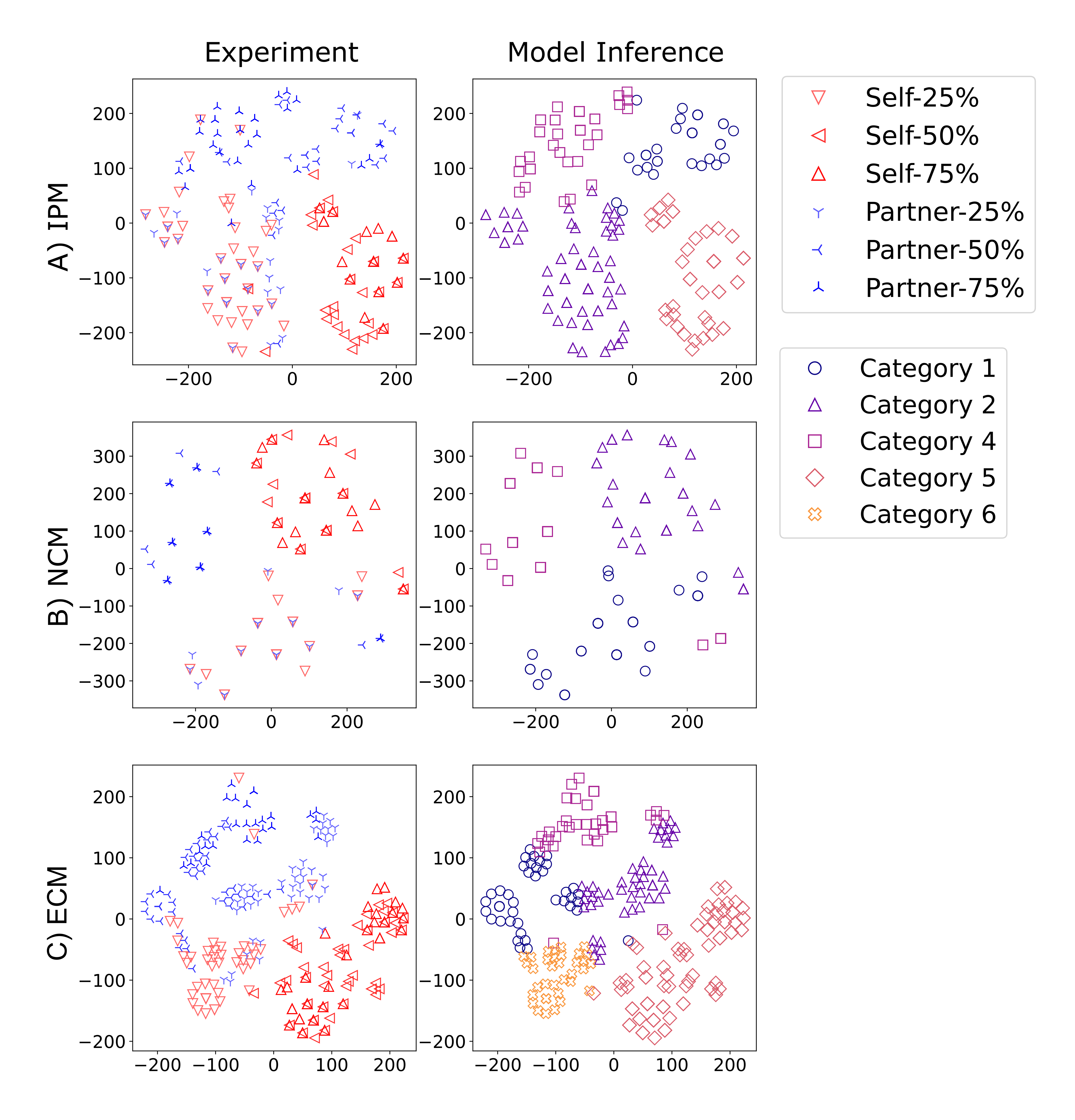}
\caption[t-SNE Visualization of Subjective-Value Representations]{
Two-dimensional t-SNE projection of probability vectors from the six-dimensional subjective-value topic distributions ($\theta$ in terms of MLDA. See Appendix ~\ref{app:mlda}). Left column: plots categorized with actual experimental conditions. Right column: plots categorized with model-predicted clusters. Since five of the six categories were used across all models, the legend for `Category 3' is omitted.
A) IPM: displaying moderate clustering ability with notable confusion between Self-25\% and Partner-25\% conditions.
B) NCM: showing poor differentiation, confusing the Partner-50\% and Partner-75\% conditions, among others.
C) ECM: achieving the most refined separation of experimental conditions, with minimal overlap and only slight confusion between the Self-50\% and Self-75\% conditions.
Note that the absolute values on the x- and y-axes in t-SNE are meaningless~\cite{JMLR:v9:vandermaaten08a}; 
only the distances between pairs of nearby points are essential for knowing the similarity of data points.
}
\label{fig:tsne_visualization}
\end{figure*}

To visually compare clustering patterns across models, test-trial representations were projected into a shared two-dimensional space using t-SNE~\cite{JMLR:v9:vandermaaten08a}(Figure~\ref{fig:tsne_visualization}). It revealed that the ECM showed a clear separation of conditions, utilizing five of the six latent topics, and even distinguished between similar Self-25\% and Partner-25\% conditions as distinct clusters. This model showed only minor confusion appeared between the Self-50\% and Self-75\% conditions, demonstrating the model's ability to represent social context as subjective value (Figure~\ref{fig:tsne_visualization}, bottom).
The IPM displayed an intermediate clustering pattern, utilizing four of the six available topics. This model combined the Self-25\% and Partner-25\% conditions into a single mixed cluster.
Additionally, it showed some confusion between the Self-50\% and Self-75\% conditions, although other high-contrast conditions remained reasonably well separated (Figure~\ref{fig:tsne_visualization} top).
The NCM exhibited the weakest structure, employing only three of the six topics. The resulting clusters showed significant overlap, with multiple experimental conditions grouped. In particular, the model failed to differentiate between the Partner-50\% and Partner-75\% conditions, and also showed confusion in other condition pairs. This confirmed the model's limited ability to incorporate social comparison information into its subjective-value representations (Figure~\ref{fig:tsne_visualization}, middle).

These qualitative patterns demonstrated that explicit access to partner-reward information (as in the ECM) produced the most refined representations of subjective value. In contrast, the IPM and NCM failed to distinguish between behaviorally relevant conditions for the monkeys. The effectiveness of topic utilization (ECM: 5/6 topics, IPM: 4/6 topics, NCM: 3/6 topics) directly correlated with each model's ability to discriminate between experimental conditions. 

\subsection{Prediction of Licking Behavior}

\subsubsection{Prediction from stimulus images}
\begin{figure*}[t]
    \centering
    \includegraphics[width=0.8\textwidth]{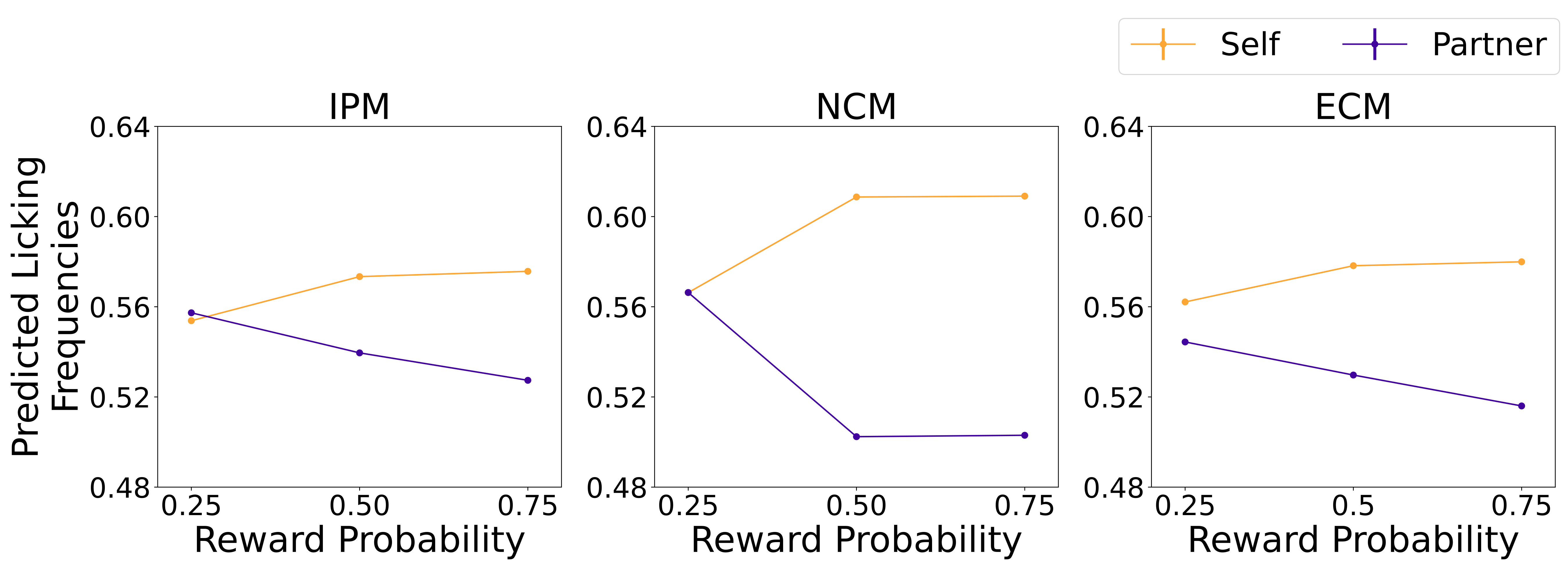}
    \caption[Predicted Licking Behavior Across Experimental Conditions]{
        Comparison of predicted licking frequencies for the self monkey across the six experimental conditions. The x-axis: three reward probability conditions; three self-variable conditions (Self-25\%, Self-50\%, and Self-75\%) and partner-variable conditions (Partner-25\%, Partner-50\%, and Partner-75\%). The y-axis: normalized licking frequency. Results for three models: IPM (left), NCM (center), and ECM (right), with ECM best preserving the experimentally observed pattern(in Figure~\ref{fig:experimental_settings}C). Center and error bars indicate mean $\pm$ s.e.m.
    }
    \label{fig:image_prediction}
\end{figure*}

We evaluated the models' ability to predict self-licking frequency across the six experimental conditions from only the image input. 
The ECM successfully captured the key behavioral patterns observed in the original experiment (Figure~\ref{fig:experimental_settings}C)~\cite{Noritake2018-hm}. In the self-variable blocks, the model accurately predicted an increase in licking frequency as the self-reward probability increased from 25\% to 75\%, reflecting the monkeys' natural anticipatory responses to more probable rewards. More critically, in the partner-variable blocks, the ECM accurately captured the decrease in licking frequency with increasing partner-reward probability, while the self monkey's reward probability remained constant at 20\%. This pattern was a defining characteristic of social comparison, showing that subjective-reward value diminishes when a partner receives greater rewards, with one's own reward remaining unchanged (Figure~\ref{fig:image_prediction} right).
The IPM, despite incorporating information about the partner's subjective state, failed to accurately capture these behavioral patterns. Most notably, this model predicted nearly identical licking frequencies for the Self-25\% and Partner-25\% conditions, despite these representing fundamentally different social contexts (Figure~\ref{fig:image_prediction}, left).
The NCM performed particularly poorly under partner-variable conditions, showing a reversed trend: licking frequency increased as partner-reward probability rose from Partner-50\% to Partner-75\%. This directly contradicted the observed behavioral pattern where monkeys displayed decreased anticipatory licking when the partner monkey received greater rewards, indicating the necessity of the partner-reward information in social comparison (Figure~\ref{fig:image_prediction}, center). 

Based on our findings, we concluded that the ECM provided the most accurate representation of the behavioral and neural processes underlying social comparison. The direct incorporation of observed partner-reward information, without the added complexity of inferring subjective states, appeared optimal for predicting social-comparison behavior.

\subsubsection{Prediction from unobserved reward distributions}
\label{subsubsec:extrapolation}

\begin{figure*}[t]
    \centering
    \includegraphics[width=0.8\textwidth]{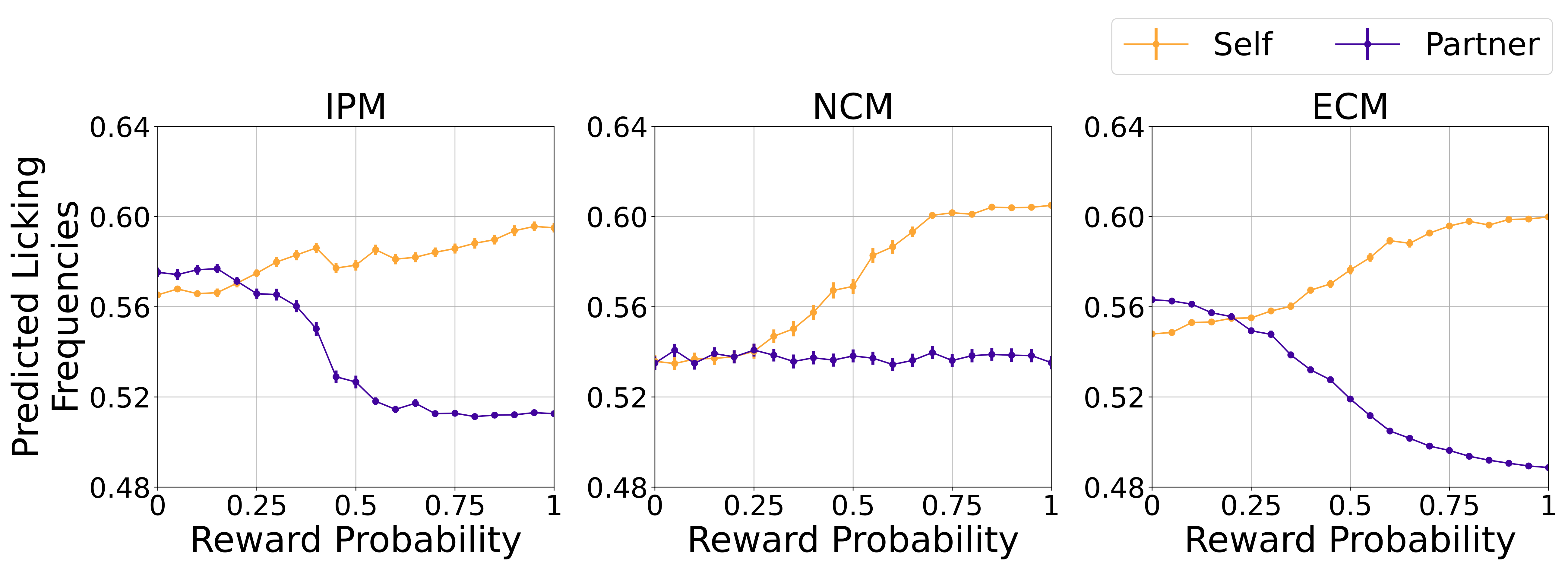}
    \caption[Predicted Licking Behavior from Unobserved Reward Distributions]{
    Predicted licking frequencies for reward distributions not present in training data. Yellow: the self-variable block (partner reward fixed at 20\%) with self-reward varying from 0--100\%. Purple: the partner-variable block (self-reward fixed at 20\%) with partner reward ranging from 0--100\%. Across the models (IPM, NCM, and ECM), ECM produced the most theoretically valid results. Center and error bars indicate mean $\pm$ s.e.m.
    }
    \label{fig:imaginary_reward_prediction}
\end{figure*}

This experiment examined how the three models predicted licking frequencies when presented with reward distributions not encountered during training, such as a self-reward probability of 35\%. Similar to the original experimental design~\cite{Noritake2018-hm}, the partner's reward probability in the self-variable block was fixed at 20\%, while the self-reward probability varied from 0\% to 100\% in 5\% increments. Conversely, the self-reward probability in the partner-variable block was fixed at 20\%, and the partner's reward probability varied from 0\% to 100\%. Each model predicted the self monkey's licking frequency after simulating 100 trials for each of the 40 conditions.
Both the IPM and the ECM successfully reproduced the key behavioral trends: increased licking frequency with increasing self-reward probability and decreased licking frequency with increasing partner-reward probability. 
However, the IPM's downward trend in the partner variable blocks leveled off above 50\%, while the ECM decreased licking frequency as the partner's probability of the reward increased. This demonstrates ECM's superiority, as the subjective value should diminish in proportion to the partner's reward probability increase.
Additionally, as expected given its architecture, the NCM showed no sensitivity to changes in partner reward, maintaining a consistent licking frequency regardless of variations in partner reward. These findings further support the superiority of the ECM (Figure~\ref{fig:imaginary_reward_prediction}).

These results indicate that the ECM not only fits the existing experimental data but also acquires robust internal representations capable of making theoretically consistent predictions for unobserved situations. Taken together, these findings suggest that direct comparisons of external rewards, rather than complex inferences about internal states, played an essential role in social comparison.

\subsection{Inter-Node Dependency Analysis Using Normalized Mutual Information}
\label{subsec:nmi}
The consistent superiority of the ECM prompted a closer examination of how the self-subjective-value node contained the information from another node. While the superiority suggests that the decrease in parameter size led to the efficient processing of partner information, we do not have a way to reveal the causal effect of knocking out the partner inference node. However, we can still analyze how the lack of inference about partner information affected the subjective value of the self. 
To address this point, we computed the normalized mutual information (NMI) between that node and every other latent or observation node. Because the NCM had no dedicated partner-reward node, the analysis compared only the IPM and ECM.
Document-based NMI values from the two models formed paired samples, which we compared using a two-sided Wilcoxon signed-rank test for every node pair \((\ast,z^{Rs})\).  
The median NMI, interquartile range (IQR), and corresponding \(Z\) and \(p\) values are reported in Table~\ref{tab:nmi_results}. Except for the pair \((w^{Rs},z^{Rs})\), NMI was significantly higher in the ECM for every node examined.  
Granting the model direct access to partner-related variables, therefore, strengthened not only the information pathway from those variables to the self-valuation node but also dependencies involving other nodes, including the nodes for the self---monkey and the image stimuli---although the ECM has fewer free parameters.

\begin{table*}[t]
    \centering
    \caption{Normalized mutual information between each node and the self-subjective-value topic \(z^{Rs}\) across 59 test days.  
     IQR follows medians in brackets.  
     Asterisks denote significance level in a two-sided Wilcoxon signed-rank test: *$p < 0.05$, **$p < 0.01$, ***$p < 0.001$.}
    \resizebox{0.8\textwidth}{!}{
        \begin{tabular}{lcccccc} 
            \toprule 
            Node & IPM Median & IPM IQR & ECM Median & ECM IQR & Z & p \\ 
            \midrule 
            $w^{Ap}$ & $6.06 \times 10^{-6}$ & $1.06 \times 10^{-5}$ & $1.40 \times 10^{-5}$ & $6.85 \times 10^{-5}$ & $-8.91$ & $5.33 \times 10^{-19***}$ \\ 
            $w^{As}$ & $6.23 \times 10^{-4}$ & $4.40 \times 10^{-4}$ & $8.18 \times 10^{-4}$ & $4.78 \times 10^{-4}$ & $-8.80$ & $1.33 \times 10^{-18***}$ \\ 
            $w^{Rp}$ & $3.66 \times 10^{-3}$ & $3.02 \times 10^{-3}$ & $7.28 \times 10^{-3}$ & $2.78 \times 10^{-2}$ & $-10.0$ & $1.04 \times 10^{-23***}$ \\ 
            $w^{Rs}$ & $1.12 \times 10^{-2}$ & $3.81 \times 10^{-3}$ & $1.08 \times 10^{-2}$ & $6.32 \times 10^{-3}$ & $-1.18$ & $2.37 \times 10^{-1}$ \\ 
            $w^{S}$ & $2.35 \times 10^{-2}$ & $1.23 \times 10^{-2}$ & $3.35 \times 10^{-2}$ & $1.51 \times 10^{-1}$ & $-7.70$ & $1.36 \times 10^{-14***}$ \\ 
            $z^{Ap}$ & $2.85 \times 10^{-1}$ & $1.36 \times 10^{-1}$ & $3.91 \times 10^{-1}$ & $5.00 \times 10^{-1}$ & $-8.94$ & $4.06 \times 10^{-19***}$ \\ 
            $z^{As}$ & $8.49 \times 10^{-1}$ & $3.10 \times 10^{-2}$ & $9.27 \times 10^{-1}$ & $6.20 \times 10^{-2}$ & $-15.2$ & $3.21 \times 10^{-52***}$ \\ 
            $z^{S}$ & $3.09 \times 10^{-1}$ & $1.44 \times 10^{-1}$ & $6.98 \times 10^{-1}$ & $3.32 \times 10^{-1}$ & $-15.3$ & $1.26 \times 10^{-52***}$ \\ 
            \bottomrule 
        \end{tabular}
    }
    \label{tab:nmi_results}
\end{table*}

%% file: sections/discussion.tex
\section{Discussion}
\subsection{Superiority of the External Comparison Model}
In this study, we employed a constructive approach to investigate whether social comparison in monkeys is influenced by either direct observation of partners' rewards or inferences about partners' subjective values. The contribution of this work is primarily that it is the first to test the functionality of partner value inference in the social comparison settings. Furthermore, our model comparison results under the current settings suggest that monkeys' social comparison behavior can be explained by direct observation of partners' rewards, without requiring explicit inference of partners' subjective values.

We assessed how well each model's subjective-value classifications aligned with the experimental conditions and how accurately licking-behavior predictions matched observed licking behavior.
The ECM achieved the highest classification accuracy and showed the greatest similarity to actual monkey behavior when the expected behavioral output was derived from stimulus images. It showed upward or downward trends according to the probabilities of increases in self-reward or partner-reward, respectively, and also distinguished between the similar Self-25\% and Partner-25\% conditions.
It also yielded the most valid predictions when the unknown reward distributions were used as inputs, particularly showing a continued decrease in the partner variable blocks above 50\%. This indicated that the model that performed social comparison based on objective partner-reward information was more consistent with actual monkey cognition and behavior than the IPM that included nodes for inferring the partner's subjective reward value.

The conclusion that monkeys may not rely on partner subjective-value inference appears to contradict recent findings suggesting that monkeys possess a theory of mind~\cite{Hayashi2020-qk, Drayton2016-yv}. However, since previous research on theory of mind has demonstrated inferential abilities regarding partners' subjective belief states, it is reasonable to consider that reward-value inference in social comparison is distinct from belief inference in theory of mind.

The implications of our results, which might suggest a lack of the partner's subjective value inference in our experimental paradigm and modeling strategy, become clearer when contrasted with other computational works that have addressed partner-valuation inference. First, our work is currently the only one to incorporate the inference of the partner's subjective valuation in social comparison settings. Second, prior works that engaged in partner valuation inference~\cite{devaine2017reading, Zaki2016-vr}, in settings distinct from social comparison, used reinforcement-learning models because subjects could maximize their rewards by selecting actions. However, the experiment by Noritake et al.~\cite{Noritake2018-hm} differs from theirs in that the visual stimuli indicated the forthcoming reward structure, and the monkeys' actions did not alter the results. The situation in which two monkeys perceived the visual stimuli and the distribution of rewards between them was more like knowing other minds through joint attention to those stimuli. Therefore, we employed a multimodal categorization strategy in our social comparison modeling that also incorporated partner valuation inference. It is possible that datasets with feedback loops might produce different results from ours.

In addition, this study quantitatively elucidated why bypassing partner subjective-value inference yielded better results by calculating normalized mutual information. The ECM showed significantly higher normalized mutual information between objective reward information and the self's subjective value, indicating that the information from the partner's rewards was transmitted more efficiently and with greater fidelity to the self-valuation node.
We propose that this was because the ECM provided a more direct pathway for partner-reward information to influence the self's subjective value representation. By removing the partner's subjective-value latent node, the model enabled precise partner-reward probabilities to directly inform the situation-awareness node, which, in turn, influenced the self's subjective-value categorization.
In contrast, the IPM introduced potential inaccuracies in inferring the partner's subjective values. The partner's licking behavior contained inherent variability and noise, which could lead to imperfect categorization in the partner's action-intention node. These inaccuracies propagated to the partner's subjective-value node, potentially degrading the quality of information available to the situation-awareness node.

However, in this analysis, the ECM showed higher normalized mutual information with the self-monkey's subjective-value node than the IPM did for almost all nodes, not just the objective reward node. The reasons for this finding require future consideration. One candidate is the effect of the `weight' parameter, as determined by the method described in Appendix~\ref{app:hyper}. It is also worth analyzing how mMLDA's model structures, including their learnable parameter size, are generally related to normalized mutual information.

\subsection{Merit of Constructive Approach}
Beyond the model comparison, the overall merit of our constructive approach lies in our generative model's ability to simulate self-monkeys' behavior and cognition, and to predict partner information, even across different social contexts. The ability was illustrated in the extrapolation experiment~\ref{subsubsec:extrapolation}. It means our models could be extended to more complex social comparison simulations in the future, whereas currently the model can still simulate social comparison in a limited setting. Although there is a social-comparison study of LLMs~\cite{Ramamoorthy2026-da} with simulation capabilities, it is not biologically informed. Thus, our model extension could lead to a more realistic simulation of social comparison in humans or animals. The simulation capability was lacking in analytic research on social comparison (e.g.,~\cite{Noritake2018-hm, Noritake2020-os, Tai2012-pk}), because of its focus on identifying necessary conditions.

With regard to simulation capability, the ECM and the IPM are superior to the NCM. Since the NCM does not learn from partner information, it cannot predict it.

\subsection{Limitation}
\label{diss:limitation}
Our work is limited in two ways: first, it is uncertain whether our modeling consistently yields the tendency observed in our results across essentially different social comparison datasets, as in those of Noritake et al.~\cite{Noritake2018-hm}. Our modeling with mMLDA, a multimodal categorization model that leverages self- and partner information, might be well-suited to the dataset from Noritake et al.~\cite{Noritake2018-hm}, as it induces social comparison during joint attention with the partner. However, social comparison across datasets might be better captured by another model, and mMLDA might exhibit different behavior. For example, with datasets that contain feedback loops, reinforcement learning might be better suited to assess whether monkeys infer the partner's subjective valuation. In fact, other works on partner valuation inference used datasets with feedback loops and thus adopted reinforcement learning, although their works were not related to social comparison~\cite{Zaki2016-vr,devaine2017reading}. 
Second, the dataset from Noritake et al.~\cite{Noritake2018-hm} might not strongly incentivize partner subjective-value inference, because the water reward itself had essentially the same value for the two monkeys. If the experiments forced the self and the partner to assign completely different subjective values to the same thing, subjective value inference might have been prompted. This can be realized, for example, by setting different exchange ratios of the items between the two subjects, which were introduced in the studies of fairness using `ultimatum games'~\cite {Kagel1996-gc}. Subjective value inference of the partner might be forced to appear only in such situations. 

\subsection{Future Directions}
\label{diss:future}
Our findings suggest three directions for advancing social comparison research: first, we can extend our analyses using more diverse models and datasets to assess whether our ECM excellence results generalize to different settings. Regarding modeling, our mMLDA design, which focuses on learning the co-occurrence of information at each layer, is not the only option; for example, we can also posit that situation awareness is driven by visual stimuli. In the future, contrasting our models with such a different hypothesis could yield deeper insights into social comparison. In terms of datasets, those with a feedback loop can produce distinct tendencies in social comparison, which reinforcement learning might capture better and could reveal a different nature of social comparison.

Second, correlating model components with local field potential recordings from the monkeys' brains during the experiments by Noritake et al.~\cite{Noritake2018-hm} could provide neurological validation of our computational findings. Identifying neural signatures corresponding to the processes represented in our models would strengthen the biological plausibility of our conclusions and offer deeper insights into the neural mechanisms underlying social comparison in primates.

Finally, a natural extension would be to conduct analogous experiments with human participants to determine whether similar computational mechanisms underlie social comparison across primates. Because existing studies of social comparison have not addressed inference about partners' reward value, it is unknown whether humans engage in subjective value inference about their partners. Although humans' developed cognitive abilities may enable inference of subjective value, it is also possible that humans possess a direct external-comparison mechanism, as found in monkeys, to maintain computational efficiency.

Furthermore, a human-focused study could incorporate additional measures, such as explicit subjective value ratings and questionnaires assessing participants' inferences about their partners' experiences. By adding measures, we could gain insight into the relationships between social comparison and other cognitive phenomena. Particularly, envy---an affective response elicited when a partner receives greater rewards, potentially serving functions in group cohesion---may modulate subjective-value representations, as supported by neuroimaging findings showing envy-driven changes in valuation circuits~\cite{Takahashi2009-ss, Dvash2010-nj, Dai2024-jg}. Additional measures would allow for direct testing of whether humans engage in partner-subjective-state inference during social comparison or primarily rely on objective reward observation.

In future human experiments, we could introduce an additional experimental setting to prompt partner subjective value inference, thereby overcoming the potential limitation of the data from Noritake et al.~\cite{Noritake2018-hm}, which did not strongly incentivize partner value inference, and incentivization could be realized by, for example, setting the different exchange ratio between subjects as discussed in the last subsection~\ref{diss:limitation}. Human experiments could incorporate such a complex setting, which is difficult with monkeys.

%% file: sections/conclusion.tex
\section{Conclusion}
This study investigated whether social comparison in a monkey requires inferring its partner's subjective values or relies solely on observing objective rewards, based on the findings of Noritake et al.~\cite{Noritake2018-hm}. We developed three computational models with varying degrees of social information processing: an IPM that infers the partner's subjective values, an NCM that disregards partner information entirely, and an ECM that directly incorporates the partner's objective rewards. Using mMLDA, we evaluated these models based on their ability to classify subjective values across six experimental conditions.

Our results under the current experimental and modeling settings suggest that social comparison in monkeys does not necessarily require inferring their partner's subjective valuations. The ECM, which directly incorporates partner-reward information without inferring subjective states, demonstrated the highest classification accuracy. This finding indicates that monkeys may primarily rely on external, objective reward information when engaging in social comparison, rather than attempting to model their partner's internal subjective experience of rewards, although the generalization should be tested in other experimental and modeling settings.

Future research correlating model components with local field potential recordings from monkeys' brains during the experiments could provide neurological validation of our computational findings, potentially illuminating the neural mechanisms underlying social comparison processes in primates. Additionally, the future study could extend this work by using non-invasive methodologies with human subjects, providing deeper insights into the nature of social comparison.

%% file: sections/appendix.tex
\appendix
\section{Technical Details of MLDA and Implementation}
\subsection{Multimodal Latent Dirichlet Allocation (MLDA)}
\label{app:mlda}

\begin{figure*}[t]
    \centering
    \includegraphics[width=0.5\linewidth]{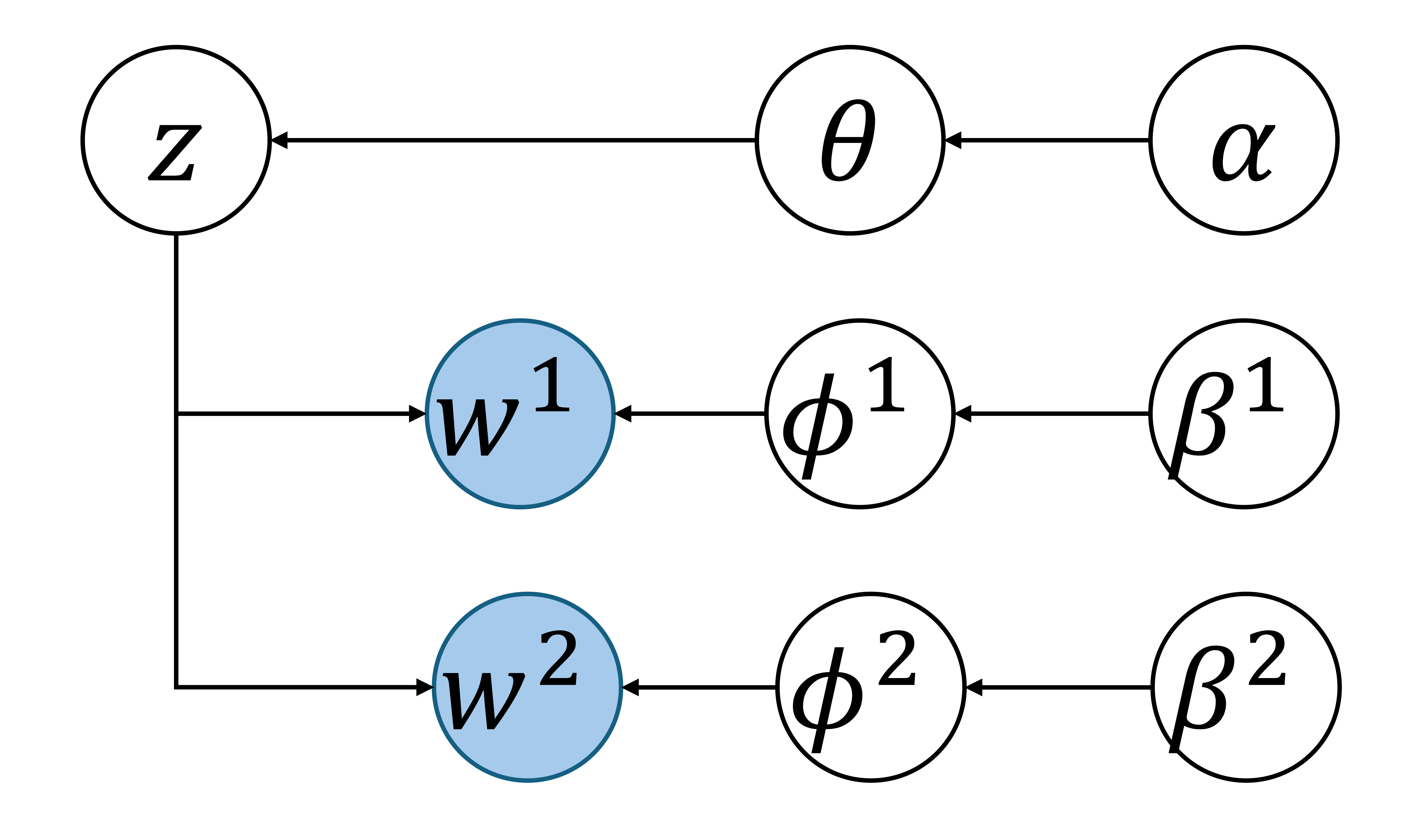}
    \caption{Graphical model representation of a general MLDA with two modalities (e.g., reward and licking). Shaded nodes: observed variables. Unshaded nodes: latent variables and parameters.}
    \label{fig:mlda}
\end{figure*}

MLDA extends the original LDA to enable category formation across multiple information types (modalities), as described in Figure \ref{fig:mlda}. The variable $z$ denotes the common latent category derived from observed data $w^m \in \{w^1, w^2\}$, which are generated through multinomial distributions parameterized by $\theta$ and $\phi^m \in \{\phi^1, \phi^2\}$.

The generative process is defined as:
\begin{align}
    z &\sim \text{Mult}(\theta), \label{eq:z_mult}\\
    w^m &\sim \text{Mult}(\phi^m). \label{eq:w_mult}
\end{align}

The Dirichlet priors governing $\theta$ and $\phi^m$ are defined by hyperparameters $\alpha$ and $\beta^m \in \{\beta^1, \beta^2\}$:
\begin{align}
    \theta &\sim \text{Dir}(\alpha), \label{eq:theta_dir}\\
    \phi^m &\sim \text{Dir}(\beta^m). \label{eq:phi_dir}
\end{align}

The learning process establishes categories by inferring $z$ from the observed data $w^m$ while optimizing parameters $\theta$ and $\phi^m$. We employed Gibbs sampling~\cite{Griffiths2004-vc} for $z$ estimation. This iterative approach updates parameters using:
\begin{align}
    &P(z_{mij} = k|\mathbf{W}, \mathbf{Z}^{\setminus mij}, \alpha, \beta^m) \nonumber \\
    &\propto (n^{\setminus mij}_{k,j} + \alpha) \Bigg(\frac{n^{\setminus mij}_{m,w^m,k} + \beta^m}
    {n^{\setminus mij}_{m,k} + W^m\beta^m} \Bigg). \label{eq:gibbs_sampling}
\end{align}

In this formulation, $\mathbf{W}$ encompasses all observations across modalities, with $J$ total observations yielding $J$ category assignments in $\mathbf{Z}$. The observations span $M$ modalities, with $W^m$ representing the vocabulary size (number of distinct features) in the $m$-th modality. The notation $\setminus$ indicates exclusion from a set, while $n_{\ast}$ denotes count statistics. Specifically, $n_{m,w^m,k,j}$ counts instances where category $k$ is assigned to observation $w^m$ in modality $m$ during the $j$-th trial. $\mathbf{Z}^{\setminus mij}$ contains all category assignments except $z_{mij}$, which corresponds to the $i$-th observation of modality $m$ in trial $j$. Without specifying $j$, the frequencies cover the entire dataset. The terms $n^{\setminus mij}_{k,j}$, $n^{\setminus mij}_{m,w^m,k}$, and $n^{\setminus mij}_{m,k}$ exclude the current observation under consideration. They represent, respectively: the frequency of category $k$ in trial $j$, the frequency of category $k$ for feature $w^m$ in modality $m$, and the total frequency of category $k$ in modality $m$. This sampling continues until $n_{\ast}$ converges, after which $\theta$ and $\phi^m$ were calculated as:
\begin{align}
    \theta_{kj} &= \frac{n_{k,j} + \alpha}{n_j + K\alpha}, \label{eq:theta} \\
    \phi^m_{w^m,k} &= \frac{n_{m,w^m,k} + \beta^m}{n_{m,k} + W^m\beta^m}, \label{eq:phi}
\end{align}
Here, $K$ represents the total number of categories.

Once trained, the model is able to categorize previously unseen data. Given observed information $\mathbf{w}_{obs}$ across modalities, the category $\hat{z}$ is estimated as:
\begin{align}
\hat{z} \sim P(z|\mathbf{w}_{obs}) = \int P(z|\theta)P(\theta|\mathbf{w}_{obs})d\theta. \label{eq:z_obs}
\end{align}

The optimal category assignment is obtained by maximizing this probability:
\begin{align}
k = \argmax_z P(z|\mathbf{w}_{obs}). \label{eq:k_obs}
\end{align}
The model is also able to predict unobserved modality information $w$:
\begin{align}
    P(w|\mathbf{w}_{obs}) &= \sum_z P(w|z)P(z|\mathbf{w}_{obs}) \nonumber \\
    &= \int \sum_z P(w|z)P(z|\theta)P(\theta|\mathbf{w}_{obs})d\theta. \label{eq:w_obs}
\end{align}
The posterior $P(\theta|\mathbf{w}_{obs})$ in equations \eqref{eq:z_obs} and \eqref{eq:w_obs} is derived via the previously described Gibbs sampling procedure. Through this approach, MLDA generates categorical representations from multimodal data in an unsupervised manner.

\subsection{Mathematical Details of the Bayesian Modeling}
\label{app:math_bayesian}

\begin{figure*}[t]
    \centering
    \includegraphics[width=0.9\textwidth]{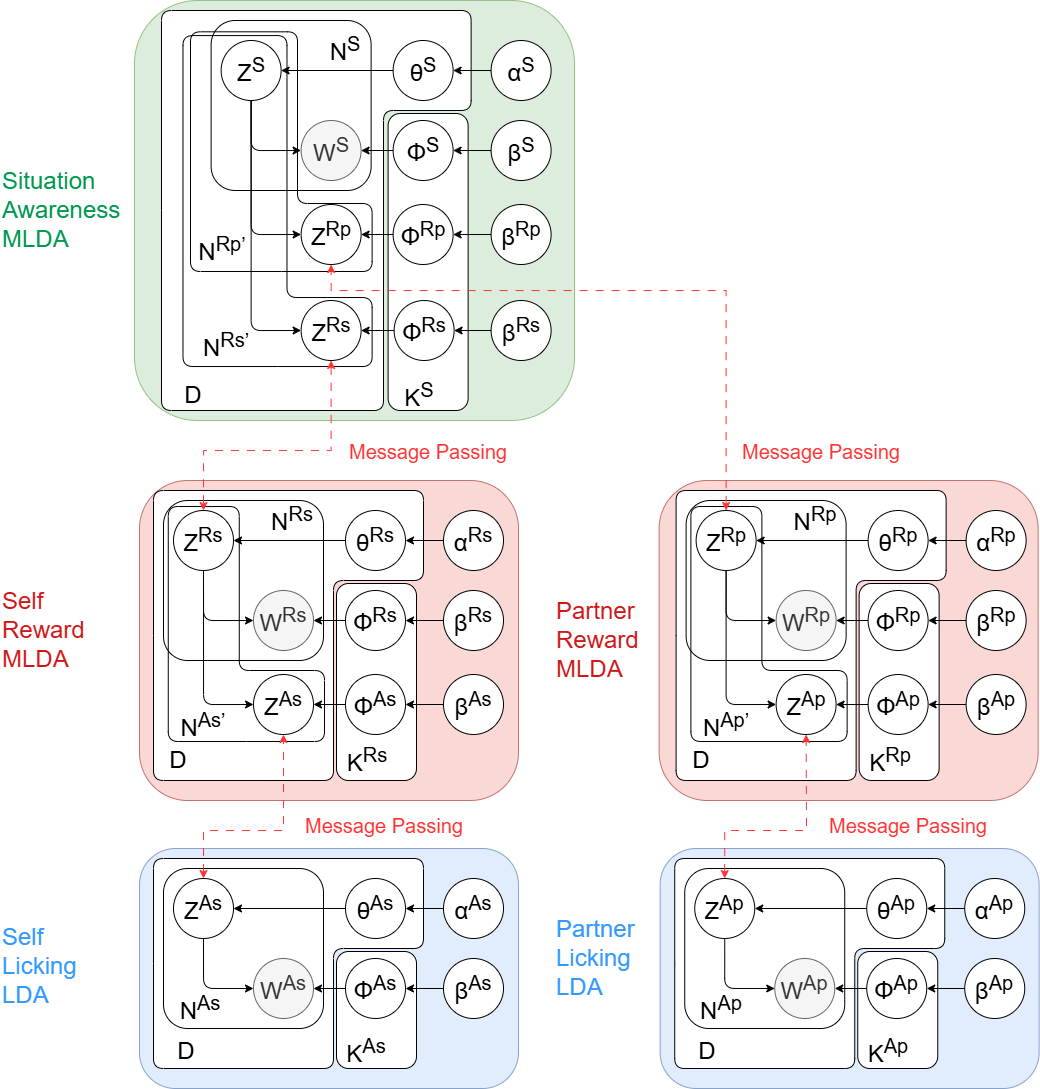}
    \caption{The exact graphical model of IPM}
    \label{fig:exact_ipm}
\end{figure*}

The mathematically exact representation of the IPM is shown in Figure~\ref{fig:exact_ipm}. Note that this figure corresponds to the simplified version of the IPM illustration shown in Figure~\ref{fig:model_architectures}A. As to the notations, $D$ is the number of data or experimental blocks in experiments by Noritake et al.~\cite{Noritake2018-hm}. Each block was converted into a multimodal `document' in LDA terminology in the way described in Appendix~\ref{app:feature}. And $D$ was shared by the entire model. Because the training had 233 days and the test 59 days, and each day had 6 experimental blocks, the $D$ for training was $1398 = 233 * 6$ and the $D$ for testing was $354 = 59 * 6$. $K^*$ is a hyperparameter, representing the topic number of each LDA or MLDA. $\alpha^*, \beta^{*}$ are hyperparameters for Dirichlet distributions. About how these hyperparameters were determined, see Appendix~\ref{app:hyper}.  $N^{*}$ is the dimension of each feature, while this is called the vocabulary size in the LDA terminology. About how the dimension of each observation was determined, see Appendix~\ref{app:feature}. The dimension of each latent node as an observation node for an upper MLDA equals $K^{*}$ of the lower node, because all $z^{*}$ have $K^{*}$ dimensions. $\theta^{*}$ and $\phi^*$ are the learnable parameters. $\theta^*$ is the estimated probability of topics for each document.  $\phi^*$ is the estimation of the frequency of features for each topic. The notation rules are the same across the three models. The connection between two modules (LDA or MLDA) is achieved via a message-passing approach within the SERKET framework~\cite{Nakamura2018-rg}.

Our IPM employs a three-layer architecture that mirrors the common self-information processing module (described as $w^{As},z^{As},w^{Rs}, z^{Rs}$ in Figure~\ref{fig:subjective_value_model}B) to the partner information processing module, maintaining the same architecture, with an additional top layer connecting them, capturing the social comparison processes between the self monkey and the partner monkey:

\begin{itemize}
    \item \textbf{Bottom Layer}: two parallel LDA modules representing both monkeys' action intentions (licking behavior)
    \item \textbf{Middle Layer}: two parallel MLDA modules representing the subjective reward values of both monkeys
    \item \textbf{Upper Layer}: an integrative MLDA module representing the self monkey's situation awareness
\end{itemize}

In the IPM, the licking LDA module at the bottom reduces the inherent noise in biological licking data. In this model, $z^{As},z^{Ap}$ represent action-intention categories, which generate observed licking behavior $w^{As}, w^{Ap}$, respectively. These categories are generated from multinomial distributions with parameters $\theta^{As}, \theta^{Ap}$, which in turn are generated from Dirichlet distributions with hyperparameter $\alpha^{As}, \alpha^{Ap}$. We describe all distributions in this model as either multinomial or Dirichlet distributions, even when their dimensions are 2 (see Appendix~\ref{app:beta} for the reason).
At the middle layer, the latent variables $z^{Rs}, z^{Rp}$ represent categories of subjective reward value, integrating information from both the action-intention categories $z^{As},z^{Ap}$ and the observed reward probability distributions $w^{Rs},w^{Rp}$. The subjective-value node in this layer learns the probabilistic co-occurrence patterns between reward frequencies and licking frequencies. 

The inference process of subjective valuations first calculates the probability distribution of action intention categories:
\begin{align}
    \hat{z}^{As}_j \sim P(z^{As}_j|w^{As}_j)\\
    \hat{z}^{Ap}_j \sim P(z^{Ap}_j|w^{Ap}_j)
\end{align}

$w^{\ast}_j$ is the observed information of the $j$th data, and $z^{\ast}_j$ is the latent variable of the $j$th data (The maximum of $j$ is $D$).
These action categories serve as observational data for the upper layer's subjective-value categories, and the action-intention categories are updated stochastically based on the subjective value model:
\begin{align}
    P(z^{As}|w^{Rs}_j,\hat{z}^{As}_j) = \sum_{z^{Rs}} P(z^{As}|z^{Rs})P(z^{Rs}
    |w^{Rs}_j, \hat{z}^{As}_j)\\
    P(z^{Ap}|w^{Rp}_j,\hat{z}^{Ap}_j) = \sum_{z^{Rp}} P(z^{Ap}|z^{Rp})P(z^{Rp}|w^{Rp}_j, \hat{z}^{Ap}_j)
\end{align}

This probability is represented by discrete and finite parameters, which can be sent back to the bottom layer's action-intention node via message passing. 
The latent variables associated with the concept's modality information are estimated using Gibbs sampling~\cite{Griffiths2004-vc}.

\begin{align}
    z^{As}_{mij}\sim P(z^{As}|W, Z^{\backslash mij}, \alpha, \beta^m)P(z^{As}|w^{Rs}_j, \hat{z}^{As}_j)\\
    z^{Ap}_{mij}\sim P(z^{Ap}|W, Z^{\backslash mij}, \alpha, \beta^m)P(z^{Ap}|w^{Rp}_j, \hat{z}^{Ap}_j)
\end{align}
where $W$ represents all the information, and $Z^{\backslash mij}$ represents a set of latent variables, except for the latent variable assigned to the modality $m$'s $i$th information of the $j$th observation (Note that the maximum of $m$ is the number of modalities for each MLDA and the maximum of $i$ is $N^{*}$). All the latent variables were learned in a complementary manner.

After the middle layer is updated, the following equations are used to update the top layer. Initially, subjective reward value nodes are sampled, given the inference of the bottom layer:

\begin{align}
    \hat{z}^{Rs}_j \sim P(z^{Rs}_j|w^{Rs}_j,z^{As}_j)\\
    \hat{z}^{Rp}_j \sim P(z^{Rp}_j|w^{Rp}_j,z^{Ap}_j)
\end{align}

Next, inference of $z^{R*}$ in the top layer is passed downward to estimate the stochastic bias of the subjective-value latent nodes.  

\begin{align}
    P(z^{Rs}|w^S_j, \hat{z}^{Rs}_j, \hat{z}^{Rp}_j) = \sum_{z^S} P(z^{Rs}|z^S)P(z^S|w^S_j, \hat{z}^{Rs}_j, \hat{z}^{Rp}_j)\\
    P(z^{Rp}|w^S_j, \hat{z}^{Rs}_j, \hat{z}^{Rp}_j) = \sum_{z^S} P(z^{Rp}|z^S)P(z^S|w^S_j, \hat{z}^{Rs}_j, \hat{z}^{Rp}_j)
\end{align}

Gibbs sampling then updates the subjective-value node under the influence of the upper layer as follows:

\begin{align}
    z^{Rs}_{mij}\sim P(z^{Rs}|W, Z^{\backslash mij}, \alpha, \beta^m)P(z^{Rs}|w^S_j, \hat{z}^{Rs}_j, \hat{z}^{R_p}_j)\\
    z^{Rp}_{mij}\sim P(z^{Rp}|W, Z^{\backslash mij}, \alpha, \beta^m)P(z^{Rp}|w^S_j, \hat{z}^{Rs}_j, \hat{z}^{R_p}_j)
\end{align}

\subsubsection{No Comparison Model}

\begin{figure*}[t]
    \centering
    \includegraphics[width=0.45\textwidth]{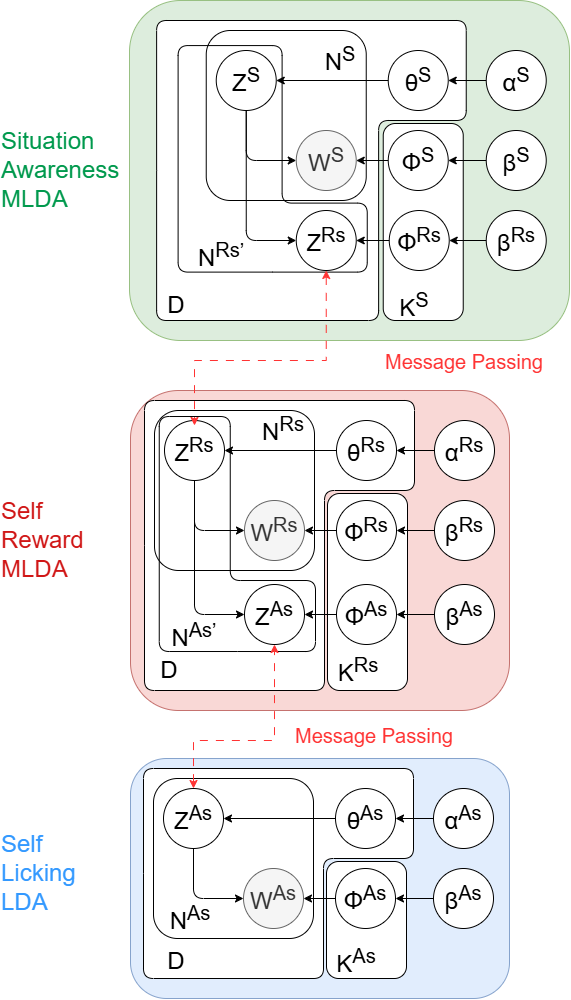}
    \caption{The exact graphical model of NCM}
    \label{fig:exact_npm}
\end{figure*}

The NCM hypothesizes that monkeys process only their own rewards without considering social or partner-related information. This model retains only the self-related components of the IPM, omitting all partner-related nodes and connections, thereby serving as a control model. The architecture includes:

\begin{itemize}
    \item \textbf{Bottom Layer}: a single LDA module for the self monkey's action intentions
    \item \textbf{Middle Layer}: a single MLDA module for the self monkey's subjective reward values
    \item \textbf{Upper Layer}: an MLDA module for situation awareness, connected only to self-related information
\end{itemize}
This architecture is shown in Figure~\ref{fig:exact_npm}, which corresponds to Figure~\ref{fig:model_architectures}B. 

The inference process follows the same hierarchical pattern with reduced components, ignoring all nodes of the partner to the information. The bottom layer is updated in the same way as the Subjective Value Model, and the middle layer is updated with the following equations:
\begin{align}
    \hat{z}^{Rs}_j \sim P(z^{Rs}_j|w^{Rs}_j, z^{As}_j)
\end{align}
\begin{align}
    P(z^{Rs}|w^S_j, \hat{z}^{Rs}_j) = \sum_{z^S} P(z^{Rs}|z^S)P(z^S|w^S_j, \hat{z}^{Rs}_j)
\end{align}
\begin{align}
    z^{Rs}_{mij}\sim P(z^{Rs}|W, Z^{\backslash mij}, \alpha, \beta^m)P(z^{Rs}|w^S_j, \hat{z}^{Rs}_j)
\end{align}
The notation and equations are the same as in the IPM, with terms restricted to self-related variables.

\subsubsection{External Comparison Model}

The ECM represents an intermediate hypothesis between the other two models. It assumes that monkeys recognize their partner's rewards but do not infer subjective valuations. The architecture includes:

\begin{itemize}
    \item \textbf{Bottom Layer}: two LDA modules (one for each monkey's action intentions)
    \item \textbf{Middle Layer}: a single MLDA module (only for the self monkey's subjective reward values)
    \item \textbf{Upper Layer}: an MLDA module for situation awareness that directly incorporates the partner monkey's observed rewards and action intentions
\end{itemize}

\begin{figure*}[t]
    \centering
    \includegraphics[width=0.9\textwidth]{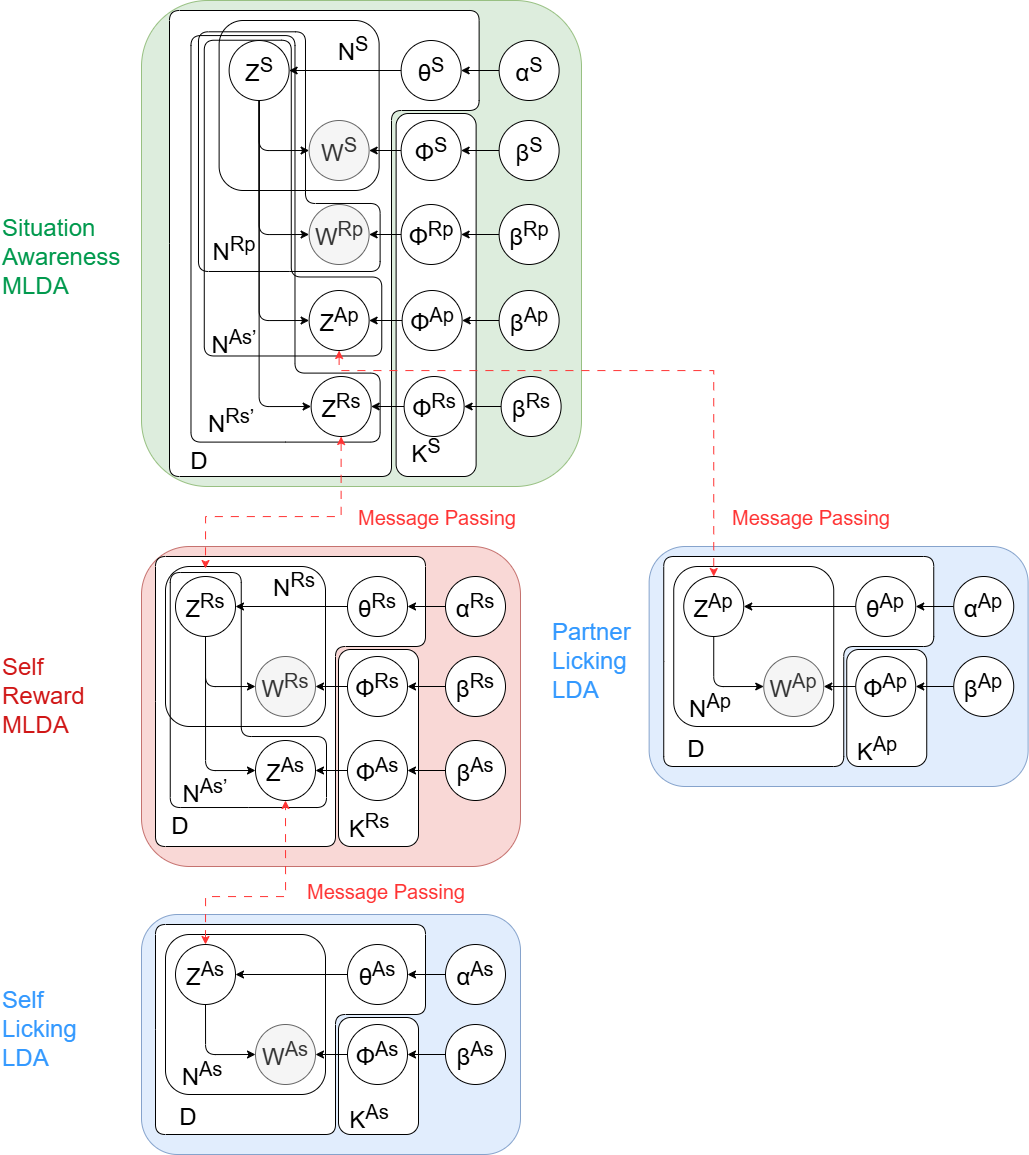}
    \caption{The exact graphical model of ECM}
    \label{fig:exact_ecm}
\end{figure*}

The critical distinction is that this model includes the partner's observed behavior ($w^{A_P}$) and reward information ($w^{R_P}$), but lacks the partner's subjective-value node ($z^{R_P}$). Instead, the partner's observable information directly connects to the situation awareness node (Figure~\ref{fig:exact_ecm}, which corresponds to Figure~\ref{fig:model_architectures}C). This model enabled us to test whether objective recognition of the partner's rewards is sufficient to explain the observed behavior, or if subjective-value inference is necessary.

The bottom layer is updated in the same way as in the other models for the self monkey, but differently for the partner monkey because the subjective valuation node for partner reward is absent. In the ECM, the inference process in the middle layer and the partner's action intention are computed using the following equations. First, samples are drawn for each latent node connected to $z^{S}$:
\begin{align}
    \hat{z}^{Rs}_j \sim P(z^{Rs}_j|w^{Rs}_j, z^{As}_j)\\
    \hat{z}^{Ap}_j \sim P(z^{Ap}_j|w^{Ap}_j)
\end{align}

Probability distributions of $z^{Rs}$ and $z^{Ap}$ are calculated using backward messages from the top layer:  
\begin{align}
    P(z^{Rs}|w^{Rp}_j, w^S_j, \hat{z}^{Rs}_j,\hat{z}^{Ap}_j) = \sum_{z^S} P(z^{Rs}|z^S)P(z^S|w^{Rp}_j, w^S_j, \hat{z}^{Rs}_j,\hat{z}^{Ap}_j)\\
    P(z^{Ap}|w^{Rp}_j, w^S_j, \hat{z}^{Rs}_j,\hat{z}^{Ap}_j) = \sum_{z^S}P(z^{Ap}|z^S)P(z^S|w^{Rp}_j, w^S_j, \hat{z}^{Rs}_j,\hat{z}^{Ap}_j)
\end{align}

Finally, the parameters of the subjective valuation node of the self monkey and the action-intention node of the partner monkey are updated in Gibbs sampling using the following two equations:
\begin{align}
    z^{Rs}_{mij}\sim P(z^{Rs}|W, Z^{\backslash mij}, \alpha, \beta^m)P(z^{Rs}|w^{Rp}_j,w^S_j, \hat{z}^{Rs}_j,\hat{z}^{Ap}_j)\\
    z^{Ap}_{mij}\sim P(z^{Ap}|W, Z^{\backslash mij}, \alpha, \beta^m)P(z^{Ap}|w^{Rp}_j,w^S_j, \hat{z}^{Rs}_j,\hat{z}^{Ap}_j)
\end{align}

\subsection{Binomial and Beta Distribution}
\label{app:beta}
Although it is terminologically accurate to describe the multinomial distribution as a binomial distribution and the Dirichlet distribution as a Beta distribution, since the licking and reward observations have two feature dimensions, we maintain the terminology of multinomial and Dirichlet distributions for consistency with the visual module and the previous literature. This is validated because when the category size of these distributions is limited
to 2, they correspond to binomial and Beta distributions, respectively.

\section{Feature Extraction from Multimodal Data}
\label{app:feature}
The raw multimodal dataset was processed into feature vectors suitable for our probabilistic generative models, which require discrete frequency vectors as input.
\subsection{Experimental Conditions and Visual Stimuli}
\label{app:stimuli}
The six distinct visual stimuli associated with the experimental conditions were processed into frequency vectors using a vector-quantized variational autoencoder~\cite{van-den-Oord2017-bw}. The model was pre-trained on the CIFAR-10 dataset~\cite{cifar} to extract low-dimensional representations of the stimuli. Each image was compared with 512 codebook vectors, selecting the most similar one based on the minimum distance.
The resulting feature representation for each image is a 512-dimensional frequency vector indicating the occurrence count of each selected codebook vector. This approach achieved an average mean squared error (MSE) of 0.0256, indicating a high-quality reconstruction of the original stimuli.

\subsection{Reward Information Processing}
\label{app:reward}
For our computational modeling, reward information was encoded as the distribution of reward outcomes within each experimental block: feature vectors were defined as $w^{R} = \{n_{reward}, n_{no\_reward}\}$, representing the count of rewarded and unrewarded trials within each block. For example, in a self-variable block with the self's reward probability at 75\% (40 trials), the feature vector is $w^{R_S} = \{30, 10\}$ for the self, indicating 30 rewarded trials and 10 unrewarded trials. Simultaneously, the partner with a fixed 20\% reward probability is represented by $w^{R_P} = \{8, 32\}$. In a partner-variable block where the partner's reward probability is 25\% (40 trials), the encoding is $w^{R_P} = \{10, 30\}$ for the partner and $w^{R_S} = \{8, 32\}$ for the self, with a fixed 20\% probability.
This representation preserved both the absolute frequency and relative proportion of rewards, allowing our models to capture these aspects of the reward distribution. Importantly, these reward distributions created social comparison scenarios in which one monkey systematically received more or fewer rewards than its partner, enabling us to examine how subjective reward valuation was influenced by social context.

One drawback of this preprocessing of reward information should be noted: an aspect of the actual experimental setting in Noritake et al.~\cite{Noritake2018-hm} was abstracted. In the experimental paradigm, each trial never provided the reward to both monkeys simultaneously.  In other words, there were only three cases: the self-monkey rewarded, the partner monkey rewarded, and neither of them rewarded. This design is intended to reflect the resource limitations in the natural environment: the other reward means that you will not receive any reward. However, since our dataset transformed trials into the variable-block-based reward frequency, it ignores the trial-based pairwise correspondence between the self and partner monkey's rewards. Therefore, we cannot determine from the reward vectors whether the original experiment involved both cases being rewarded. Still, the resulting vectors held the information of reward imbalance, enabling our successful analysis in the subsequent sections.

\subsection{Licking Behavior Processing}
\label{app:licking}
The original licking data consisted of binary time 
series (0 or 1). The high-dimensional temporal licking data were processed as follows: the average licking frequency was calculated across all trials in the block within each selected time window (401--800 ms after stimulus onset for self in the self-variable blocks, 401--700 ms after stimulus onset for partner in the self-variable blocks, and 701--1,000 ms after stimulus onset for both monkeys in the partner-variable blocks). Data normalization was performed per day using z-score standardization across the six experimental blocks to account for day-to-day variability in baseline licking behavior. Outlier removal was not performed to preserve the data's natural variability. The normalized values were then scaled by a constant factor of 1,000, rounded to integers, and offset to ensure non-negativity, for compatibility with mMLDA, which takes discrete frequency data. The final feature vector for licking behavior was defined as $w^{A} = \{f_{licking}, f_{no\_licking}\}$, representing the frequency of licking and non-licking behaviors within each block.

This approach balanced dimensionality reduction with the preservation of key temporal dynamics related to anticipatory licking, which was identified as a crucial behavioral indicator of subjective reward valuation. The processed licking data captured the essential aspects of how social comparison affects reward expectation behavior, allowing our models to investigate whether monkeys infer their partner's subjective reward valuations during social comparison.

\section{Hyperparameter Determination}
\label{app:hyper}
Given that the experimental design featured six distinct reward conditions, the number of categories in each latent node was set to $K^{A_S} = K^{A_P} = K^{R_S} = K^{R_P} = K^S = 6$, corresponding to the six distinct experimental conditions. This configuration allowed our models to capture the full range of experimental situations. For the Dirichlet priors, uniform values were assigned across all latent variables:
\begin{align}
\alpha^{A_S} = \alpha^{A_P} = \alpha^{R_S} = \alpha^{R_P} = \alpha^S = 1.0
\end{align}
\begin{align}
\beta^{A_S} = \beta^{A_P} = \beta^{R_S} = \beta^{R_P} = \beta^S = 1.0
\end{align}

For the remaining parameters (`weights'), Bayesian optimization was used to tune the modality-specific weights. This approach systematically searched the hyperparameter space to minimize the Kullback-Leibler (KL) divergence between predicted and actual self-licking behavior frequencies on held-out test data. Likelihood-based metrics were not used because adjusting weights in MLDA alters the sample sizes in multinomial distributions in the SERKET framework~\cite{Nakamura2018-rg} as `weights' scale feature vectors. This modification of sample sizes changes the likelihood, preventing MLDA models from being compared across different weights. KL divergence was adopted because it ignores sample sizes and enables identification of the model whose predictions most closely match the actual monkey behaviors. The objective function minimized was:
\begin{align}
D_{KL}(P_{true}||P_{pred}) = \sum_{i} P_{true}(w^{A_S}_i) \log \frac{P_{true}(w^{A_S}_i)}{P_{pred}(w^{A_S}_i|\mathbf{w}_{obs})}
\end{align}
where $P_{true}(w^{A_S}_i)$ represents the true probability distribution of the self monkey's licking behavior in test data, and $P_{pred}(w^{A_S}_i|\mathbf{w}_{obs})$ represents the predicted probability distribution of the self monkey's licking behavior, given observed information from all other modalities:
\begin{align}
\mathbf{w}_{obs} = \{w^{R_S}, w^{A_P}, w^{R_P}, w^S\}
\end{align}
Weights for the observed data were set to baseline values. These baseline values were selected based on SERKET's implementation characteristics. Small values for weights (approximately equivalent to sample sizes) led to learning failures because observation vectors were truncated to integer values, necessitating sufficiently large baseline sample sizes. Conversely, large values were unsuitable due to the computational cost of Gibbs sampling. Consequently, these specific values were chosen to achieve an appropriate balance. Concretely, licking and reward frequencies were fixed at 200, and the weight for the visual stimuli was set to 300. The larger weight of visual stimuli was due to their larger dimensions.

In the learning process, the number of iterations was fixed at 100 for each LDA or MLDA module. However, in SERKET, an mMLDA model must be updated several times to ensure correct message passing between connected modules. This process required multiple iterations, and the number of iterations was set to 3, as empirical testing confirmed that this was sufficient for the model to converge. Therefore, each module was updated 300 times, with messages from lower latent nodes changing after every 100 updates.

\section{Normalized Mutual Information}
\label{app:nmi}
For two discrete random variables \(A\) and \(B\), mutual information is defined as
\[
I(A;B)=H(A)+H(B)-H(A,B)=H(A)-H(A\!\mid\!B),
\]
where \(H(\cdot)\) denotes Shannon entropy.  
Directly comparing raw \(I(A; B)\) across models is inappropriate because the marginal entropies \(H(A)\) and \(H(B)\) differ by design.  
We therefore adopted the symmetric, scale-free normalization.
\[
\mathrm{NMI}(A,B)=\frac{2\,I(A;B)}{H(A)+H(B)},
\]
which ranges from \(0\) (independence) to \(1\) (perfect dependence) and equals the harmonic mean of the two one-sided ratios \(I/H\)~\cite{Kvalseth1987-js}.

Each model produces a full joint distribution over its nodes.  
For the IPM, the joint for one day's test data factorizes as
\begin{align*}
p_d(&w^{As},w^{Ap},w^{Rs},w^{Rp},w^{S},
      z^{As},z^{Ap},z^{Rs},z^{Rp},z^{S}) \\
  ={}& p(w^{As}\!\mid\! z^{As})\,p(z^{As}\!\mid\! z^{Rs})
       p(w^{Rs}\!\mid\! z^{Rs})\,p(z^{Rs}\!\mid\! z^{S}) \\
  &\times p(w^{Ap}\!\mid\! z^{Ap})\,p(z^{Ap}\!\mid\! z^{Rs})
         p(w^{Rp}\!\mid\! z^{Rp})\,p(z^{Rp}\!\mid\! z^{S}) \\
  &\times p(w^{S}\!\mid\! z^{S})\,p_d(z^{S}),
\end{align*}
where each conditional \(p(\cdot\!\mid\!\cdot)\) corresponds to a learned topic-word distribution \(\phi\) (or modality-specific \(\phi^m\)), and \(p_d(z^{S})\) is the document-specific topic proportion \(\theta_d\).  
The ECM factorizes analogously but with a simpler structure:
\begin{align*}
p_d(&w^{As},w^{Ap},w^{Rs},w^{Rp},w^{S},
      z^{As},z^{Ap},z^{Rs},z^{S}) \\
  ={}& p(w^{As}\!\mid\! z^{As})\,p(z^{As}\!\mid\! z^{Rs})
       p(w^{Rs}\!\mid\! z^{Rs})\,p(z^{Rs}\!\mid\! z^{S}) \\
  &\times p(w^{Ap}\!\mid\! z^{Ap})\,p(z^{Ap}\!\mid\! z^{S})
         p(w^{Rp}\!\mid\! z^{S}) \\
  &\times p(w^{S}\!\mid\! z^{S})\,p_d(z^{S}).
\end{align*}
Because the NCM has no dedicated partner-reward node, the analysis compared only the IPM and ECM.
For each of the 59 test days (composed of 6 probability blocks), we analytically marginalized these joint tables to obtain the bivariate distributions required for \(I(A; B)\) and hence NMI.

\section{Chance Level of Rand Index}
\label{app:chance_level}
The chance level of the Rand Index for our task can be calculated in closed form. Because the six experimental conditions were distributed equally (1/6) in the original experiment~\cite{Noritake2018-hm}, both clusterings (Ground Truth and predictions) can be seen to arise from uniform distributions over the six discrete variables (equal to dice) during the calculation of the chance level. Thus, given the definition of Rand index~\cite{Hubert1985-wh}, the chance level is the same as the probability of the following two cases after two ideal dice are rolled twice: 1) in both trials, the numbers that the two dice have are the same ($1/6$), and 2) in both trials, the two dice have different numbers ($5/6$). Cases where the numbers of the two dice are the same at one time, but different at the other time, are excluded.

The probability of the first case is: 
\begin{align*}
1/36 = 1/6 * 1/6
\end{align*}

And that of the second case is:
\begin{align*}
25/36 = 5/6 * 5/6
\end{align*}

So, the chance level is:
\begin{align*}
(1 + 25) /36 \simeq 0.72
\end{align*}

\section{Inferior Adjusted Rand Index of the IPM to the NCM}
\label{app:adjusted_rand_index}
To know why the IPM was inferior to the NCM in adjusted Rand index (ARI), while the opposite was true with regards to Rand index (RI), the definition of ARI and its relationship to RI are critical (for the result of RI and ARI across three models, see Table\ref{table:classification_accuracy}). The core idea of ARI is that it corrects RI for chance. When the RI equals the expected RI from two random clusterings, ARI equals $0$; on the other hand, the maximum is $1$, which occurs when two clusterings perfectly match. And if the RI is below the expected random RI, ARI can take a negative value, with the minimum of $-0.5$. Notating the random RI as $\mathbb{E}[RI]$, the relationship between ARI, RI and $\mathbb{E}[RI]$ is expressed as:

\begin{align*}
    ARI = \frac{RI - \mathbb{E}[RI]}{1 - \mathbb{E}[RI]}
\end{align*}
The definition suggests that ARI increases when RI gets higher, and $\mathbb{E}[RI]$ remains the same.

\begin{table}[h]
\centering
\caption{Contingency Matrix with Marginals (General Form)}
\begin{tabular}{c|ccc|c}
     & $Y_1$& $\cdots$ & $Y_j$& Sum ($a_i$) \\ \hline
     $X_1$& $n_{11}$ & $\cdots$ & $n_{1j}$ & $a_1$ \\
     $\vdots$ & $\vdots$ & $\ddots$ & $\vdots$ & $\vdots$ \\
     $X_i$& $n_{i1}$ & $\cdots$ & $n_{ij}$ & $a_i$ \\ \hline
     Sum ($b_j$) & $b_1$ & $\cdots$ & $b_j$ & $n$ \\
\end{tabular}
\label{table:confusion_general}
\end{table}

To calculate $\mathbb{E}[RI]$, a confusion matrix describing the pairwise co-occurrence of two clusterings is used, the general form of which is shown in Table \ref{table:confusion_general}. Here, the marginals ($a_i$, $b_j$) mean the partitions of each clustering ($X$, $Y$). Using the marginals and the total number of the dataset, three statistics are obtained to calculate $\mathbb{E}[RI]$:
\begin{align*}
    A = \sum_i \binom{a_i}{2}, B = \sum_j \binom{b_j}{2}, N = \binom{n}{2}
\end{align*}

$A$ is the sum of the data pair combinations that are calculated within each partition of $X$, and $B$ is the sum for $Y$. And $N$ is the number of all data pair combinations.

With the three statistics in hand, $\mathbb{E}[RI]$ is defined as:

\begin{align*}
    \mathbb{E}[RI] = \frac{N - A - B + 2AB/N}{N}
\end{align*}
It is worth noting that the expected RI assumes that the partitions of the two clusterings are fixed.
This means that $\mathbb{E}[RI]$ fixes the partition structure of two clusterings and then calculates the expected RI when the co-occurrence ($n_{ij}$) is completely random. Therefore, $\mathbb{E}[RI]$ is different from the chance level of RI calculated in Appendix~\ref{app:chance_level}, because the latter is a random value that ignores the partitions of the two clusterings.

\begin{table}[t] 
    \centering
    \caption{Classification Accuracy of Subjective-Value Nodes with $\mathbb{E}[RI]$}
    \begin{tabular}{lcc}
    \hline
    \textbf{Metric} & \textbf{IPM} & \textbf{NCM} \\
    \hline
    Rand Index & 0.85 & 0.83 \\
    $\mathbb{E}[RI]$ & 0.65 & 0.61 \\ 
    Adjusted Rand Index & 0.56 & 0.57 \\
    \hline
    \end{tabular}
    \label{table:ari_calculation}
\end{table}

The lower ARI of the IPM than that of the NCM resulted from the IPM's relatively lower RI given its $\mathbb{E}[RI]$, as shown in Table\ref{table:ari_calculation}. From the two confusion matrices of the two models, $\mathbb{E}[RI]$ of the IPM was calculated as $0.65$, and that of the NCM was $0.61$, meaning that the IPM's RI should be higher to be better than the NCM in ARI.

\begin{table}[t]
\centering
\caption{Contingency Matrix for the IPM}
\begin{tabular}{cc|cccccc|c}
     & & \multicolumn{6}{c|}{Cluster (Predicted)} & \\
     & & $C_1$ & $C_2$ & $C_3$ & $C_4$ & $C_5$ & $C_6$ & Sum ($a_i$) \\ \hline
    \multirow{6}{*}{\rotatebox{90}{GT}} 
     & Class 1 &  0 & 53 &  6 &  0 & 0 & 0 & 59 \\
     & Class 2 &  1 &  2 &  0 & 56 & 0 & 0 & 59 \\
     & Class 3 &  0 &  0 &  0 & 59 & 0 & 0 & 59 \\
     & Class 4 &  2 & 57 &  0 &  0 & 0 & 0 & 59 \\
     & Class 5 & 51 &  3 &  5 &  0 & 0 & 0 & 59 \\
     & Class 6 &  6 &  1 & 52 &  0 & 0 & 0 & 59 \\ \hline
     & Sum ($b_j$) & 60 & 116 & 63 & 115 & 0 & 0 & 354 \\
\end{tabular}
\label{table:contingency_ipm}
\end{table}

\begin{table}[t]
    \centering
    \caption{Contingency Matrix for the NCM}
    \begin{tabular}{cc|cccccc|c}
     & & \multicolumn{6}{c|}{Cluster (Predicted)} & \\
     & & $C_1$ & $C_2$ & $C_3$ & $C_4$ & $C_5$ & $C_6$ & sum ($a_i$)\\ \hline
    \multirow{6}{*}{\rotatebox{90}{GT}} 
     & Class 1 & 59 & 0 & 0 & 0 & 0 & 0 & 59 \\
     & Class 2 & 0 & 59 & 0 & 0 & 0 & 0 & 59 \\
     & Class 3 & 0 & 59 & 0 & 0 & 0 & 0 & 59 \\
     & Class 4 & 59 & 0 & 0 & 0 & 0 & 0 & 59 \\
     & Class 5 & 0 & 0 & 59 & 0 & 0 & 0 & 59 \\
     & Class 6 & 0 & 0 & 59 & 0 & 0 & 0 & 59 \\ \hline
     & sum ($b_j$)& 118 & 118 & 118 & 0 & 0 & 0 & 354 \\
\end{tabular}
\label{table:contingency_ncm}
\end{table}

Their confusion matrices explain why the IPM was inferior in ARI: although it used more topics than the NCM, which could lead to higher RI, the IPM's predictions were dirty, with multiple predicted clusters corresponding to a single GT class (predictions in each row scattered, such as $[0, 53, 6, 0, 0, 0]$ as, shown in Table \ref{table:contingency_ipm}). On the contrary, the NCM made pure predictions, with each predicted cluster corresponding to a single GT class (each row contained only $0$ or $59$), as shown in Table\ref{table:contingency_ncm}. The IPM's heterogeneous correspondence to GT classes led to both lower RI and higher $\mathbb{E}[RI]$, resulting in lower ARI.

\begin{table}[t]
\centering
\caption{Contingency Matrix for Best Prediction with Four Topics Usage}
\begin{tabular}{cc|cccccc|c}
     & & \multicolumn{6}{c|}{Cluster (Predicted)} & \\
     & & $C_1$ & $C_2$ & $C_3$ & $C_4$ & $C_5$ & $C_6$ & Sum ($a_i$) \\ \hline
    \multirow{6}{*}{\rotatebox{90}{GT}} 
     & Class 1 &  0 & 59 &  0 &  0 & 0 & 0 & 59 \\
     & Class 2 &  0 &  0 &  0 & 59 & 0 & 0 & 59 \\
     & Class 3 &  0 &  0 &  0 & 59 & 0 & 0 & 59 \\
     & Class 4 &  0 & 59 &  0 &  0 & 0 & 0 & 59 \\
     & Class 5 & 59 &  0 &  0 &  0 & 0 & 0 & 59 \\
     & Class 6 &  0 &  0 & 59 &  0 & 0 & 0 & 59 \\ \hline
     & Sum ($b_j$) & 59 & 118 & 59 & 118 & 0 & 0 & 354 \\
\end{tabular}
\label{table:contingency_ideal}
\end{table}

To confirm if the split of predictions really affected ARI of the IPM, it is useful to conceive an imaginary case where the confusion matrix was such that each row contained only $0$ or $59$, as shown in Table \ref{table:contingency_ideal}. In this case, the result could be calculated much better than the IPM, while maintaining roughly the same $\mathbb{E}[RI]$: $\mathbb{E}[RI]$ was $0.65$ ($0.65$ for the IPM, while smaller digits were rounded), RI was $0.89$ ($0.85$ for the IPM), and ARI was $0.68$ ($0.56$ for the IPM). In summary, while the IPM's clustering had greater potential due to more similar topic usage to the GT than the NCM's, splitting predicted clusters across a single GT class resulted in a lower ARI than the ideal prediction with the same topic usage.